\documentclass{article}
\usepackage[toc]{appendix}
\PassOptionsToPackage{square,sort,comma,numbers}{natbib} %

     \usepackage[preprint]{neurips_2019}

\usepackage[utf8]{inputenc} 
\usepackage[T1]{fontenc}    
\usepackage{hyperref}       
\usepackage{url}            
\usepackage{booktabs}       
\usepackage{amsfonts}       
\usepackage{nicefrac}       
\usepackage{microtype}      

\usepackage{epsfig}
\usepackage{graphicx}
\usepackage{amsmath}
\usepackage{amssymb}
\usepackage{bm}
\usepackage{scalerel}
\usepackage{array,multirow} 
\usepackage{graphicx}
\graphicspath{{figures/}}

\usepackage{float}
\usepackage{wrapfig} 
\usepackage{subcaption}
\usepackage{array}
\usepackage{makecell} 
\usepackage{verbatim} 
\usepackage{soul} 
\usepackage[table]{xcolor} 

\usepackage{xspace}
\newcommand{\latinphrase}[1]{\textit{#1}}         
\newcommand{\etal}{\latinphrase{et~al.}\xspace}   

\usepackage{tikz} 
\usetikzlibrary{arrows}
\usetikzlibrary{arrows.meta}
\usetikzlibrary{shapes.arrows}
\usetikzlibrary{decorations.text}
\definecolor{my_green}{rgb}{0.0, 0.9, 0.24}

\usepackage{amsmath,amsfonts,bm}

\def\eqref#1{equation~\ref{#1}}

\def\ceil#1{\lceil #1 \rceil}

\def\1{\bm{1}}

\def\vf{{\bm{f}}}

\def\vh{{\bm{h}}}

\def\vx{{\bm{x}}}
\def\vy{{\bm{y}}}

\def\evb{{b}}

\def\evg{{g}}
\def\evh{{h}}

\def\evz{{z}}

\DeclareMathAlphabet{\mathsfit}{\encodingdefault}{\sfdefault}{m}{sl}
\SetMathAlphabet{\mathsfit}{bold}{\encodingdefault}{\sfdefault}{bx}{n}

\def\sD{{\mathbb{D}}}

\def\sX{{\mathbb{X}}}

\DeclareMathOperator*{\argmax}{arg\,max}

\let\emptyset\varnothing
\newcommand{\lInfAttackD}[3]{$l_\infty$ norm-bound $\epsilon=#1$, $#2$ steps per sample, and a step size of $#3$}

\newcommand{\mlp}[2]{\textsc{mlp}{\footnotesize-$#1$$\times$$[#2]$}}

\newcommand{\advnet}{Adv}

\usepackage{amsthm}

\newtheorem{theorem}{Theorem}
\newtheorem*{theorem-non}{Theorem}
\newtheorem{defn}{Definition}

\newcommand{\dnn}{\textsc{DNN} }
\newcommand{\dnns}{\textsc{DNN}s}
\newcommand{\dnnone}{\textsc {DNN\textsubscript{1}}}
\newcommand{\dnntwo}{\textsc {DNN\textsubscript{2} }}

\newcommand{\fone}{\vf_1}
\newcommand{\ftwo}{\vf_2}
\newcommand{\eftwo}{f_2}

\usepackage{algorithm}
\usepackage{algpseudocode}
\newcommand{\AlgorithmFontSize}{\footnotesize}

\newcommand\ChangeRT[1]{\noalign{\hrule height #1}}

\newcommand{\gcell}[2]{\cellcolor{gray!15}{\tiny(#1, #2)}}
\newcommand{\gcellhead}[2]{\cellcolor{gray!15}{\tiny\textbf{(#1, #2)}}}

\title{Equivalent and Approximate Transformations of \\ Deep Neural Networks}

\author{Abhinav Kumar\\
University of Utah\\
Salt Lake City, UT\\
\texttt{abhinav.kumar@utah.edu}
\And
Thiago Serra\\
Mitsubishi Electric Research Labs\\
Cambridge, MA\\
\texttt{tserra@merl.com}
\And
Srikumar Ramalingam\\
University of Utah\\
Salt Lake City, UT\\
\texttt{srikumar@cs.utah.edu}
}

\begin{document}
    \maketitle
    
    \begin{abstract}
    Two networks are equivalent if they produce the same output for any given input. In this paper, we study the possibility of transforming a deep neural network to another network with a different number of units or layers, which can be either equivalent, a local exact approximation, or a global linear approximation of the original network. On the practical side, we show that certain rectified linear units (ReLUs) can be safely removed from a network if they are always active or inactive for any valid input. If we only need an equivalent network for a smaller domain, then more units can be removed and some layers collapsed. On the theoretical side, we constructively show that for any feed-forward ReLU network, there exists a global linear approximation to a 2-hidden-layer shallow network with a fixed number of units. This result is a balance between the increasing number of units for arbitrary approximation with a single layer and the known upper bound of $\ceil{log(n_0+1)}+1$ layers for exact representation, where $n_0$ is the input dimension. While the transformed network may require an exponential number of units to capture the activation patterns of the original network, we show that it can be made substantially smaller by only accounting for the patterns that define linear regions. Based on experiments with ReLU networks on the MNIST dataset, we found that $l_1$-regularization and adversarial training reduces the number of linear regions  significantly as the number of stable units increases due to weight sparsity. Therefore, we can also intentionally train ReLU networks to allow for effective loss-less compression and approximation.
    \end{abstract}
    
\section{Introduction}
    Deep Neural Networks (\dnns) have shown remarkable success in many domains, such as computer vision~\cite{Krizhevsky2012, Ciresan2012, Goodfellow2013,Szegedy2015,He2016DeepRL}, speech~\cite{Hinton2012}, and natural language processing~\cite{sutskever2014sequence}. While these networks show impressive results, it is not clear if these results depend on such \dnns~being as wide or as deep as they currently are. 
    In this paper, we are interested in the transformation of \dnns, especially to reduce their size or depth. 
    More generally, we aim to help answering the following question: given a network \dnnone, can we find an \emph{equivalent} network \dnntwo with a different network architecture? 
    \begin{defn}[Equivalence]
        Two deep neural networks \dnnone and \dnntwo with associated functions $\fone : \mathbb{R}^{n_0}\rightarrow \mathbb{R}^{m}$ and $\ftwo : \mathbb{R}^{n_0}\rightarrow \mathbb{R}^{m}$, respectively, are equivalent if $\fone(\vx) = \ftwo(\vx) ~\forall~\vx \in \mathbb{R}^{n_0}$.
        \label{def:equivalent_networks_1}
    \end{defn}
    In this paper we also consider networks that are not completely equivalent, but which could potentially be as good as equivalent networks, with the following divergent generalizations of Definition~\ref{def:equivalent_networks_1}:
    \begin{defn}[Local Equivalence]
        Two deep neural networks \dnnone and \dnntwo with associated functions $\fone : \mathbb{R}^{n_0}\rightarrow \mathbb{R}^{m}$ and $\ftwo : \mathbb{R}^{n_0}\rightarrow \mathbb{R}^{m}$, respectively, are local equivalent -- or locally exact -- with respect to a sub-domain $D \subseteq \mathbb{R}^{n_0}$ if $\fone(\vx) = \ftwo(\vx) ~\forall~\vx \in D$.
        \label{def:equivalent_networks_2}
    \end{defn}
    \begin{defn}[Global Linear Approximation]
        For a given deep neural network \dnnone with an associated function $\fone : \mathbb{R}^{n_0}\rightarrow \mathbb{R}^{m}$, a deep neural network \dnntwo with an associated function $\ftwo : \mathbb{R}^{n_0}\rightarrow \mathbb{R}^{m}$ is a global linear approximation of the first if, for a given norm $\ell$ and distance $\epsilon$, $\fone(\vx)=\ftwo(\vx)$ for any $\vx \in \mathbb{R}^{n_0}$ such that any input in 
        $\{ \vx' : \|\vx-\vx'\|_{\ell} \le \epsilon\}$ activates the same units in \dnnone.
    \label{def:equivalent_networks_3}
    \end{defn}
    \begin{figure}[!htb]
    \centering
    \begin{minipage}{.425\textwidth}
        \centering
        \input{figures/figure_stability.tex}
        \captionof{figure}{Equivalent transformation to a smaller network using stable units (always active or always inactive ones).}
        \label{fg.model_compression}
    \end{minipage}%
    ~~~~
    \begin{minipage}{.525\textwidth}
        \centering
        \input{figures/fig_transformation_all_activations.tex}
        \captionof{figure}{Global linear approximation to a shallow two-hidden-layer network using \emph{activation patterns}.}
        \label{fg.intro_figure}
    \end{minipage}
    \end{figure}

    One of the main practical results in this paper is that, for networks consisting of Rectifier Linear Units~(ReLUs),  
    we safely remove units that are always inactive,  combine those that are always active, and even collapse some layers in order to obtain a smaller equivalent network. On the left of Figure~\ref{fg.model_compression}, we show a \dnn with some ReLUs that are always active or inactive, i.e. stable for any $\vx \in \sD \subseteq \mathbb{R}^{n_0}$. On the right, we show an equivalent \dnn in which all inactive units (red circles) and some of the active units (green circles) are removed. Many more units can be removed if the resulting network only needs to be locally equivalent, in which case some or all the hidden layers may ultimately be collapsed. That can be used to create simpler networks to evaluate a local domain.

    One of the main theoretical results in this paper is that any \dnn with ReLUs and $L$ hidden layers has a shallow global linear approximation. On the left of Figure~\ref{fg.intro_figure}, we show a deep neural network with ReLUs and $L$ hidden layers. On the right, we show a shallow transformed network with only 2 hidden layers that globally approximates the first to arbitrary precision.
\section{Related Work}
    Our work relates reparameterization, or equivalent transformations, in graphical models~\cite{Koval1976, Wainwright2004, Werner2005}. If two parameter vectors $\theta$ and $\theta'$ define the same energy function (i.e., $E(\vx|\theta)) = E(\vx|\theta'),\forall~\vx$), then $\theta'$ is called the reparameterization of $\theta$. Reparameterization has played a key role in several inference problems such as belief propagation~\cite{Wainwright2004}, tree-weighted message passing~\cite{Wainwright2005}, and graph cuts~\cite{Kolmogorov2007}.
  
    The idea is also associated with characterizing the functions that can be represented by \dnns~\cite{Hornik1989,Cybenko1989,Telgarsky2016,Lin2018,Arora2018}. It is well known that a single, huge, hidden layer is a universal approximator of Borel measurable functions~\cite{Hornik1989,Cybenko1989}. For restricted functions such as any Booleans, we can represent it using a single-hidden layer threshold network (See ~\cite{Csaji2001} for a detailed survey of classical results on universal approximation). There has been a few results for functions more general than the Boolean ones. Recently, Lin and Jegelka~\cite{Lin2018} show that networks with deep residual layers~\cite{He2016DeepRL} and single ReLUs in every hidden layer is a universal approximator. By observing that any continuous piecewise linear function be be expressed as a difference of two convex piecewise linear functions, Goodfellow \etal~\cite{Goodfellow2013} show that the maxout network can be a universal approximator. Arora \etal~\cite{Arora2018} show that 
    every \dnn with ReLUs define piecewise linear functions and every piece-wise linear function $f : \mathbb{R}^{n_0}\rightarrow \mathbb{R}$ can be represented by a \dnn with at most $\ceil{log(n_0+1)}+1$ layers. This result is derived using the results of Wang and Sun~\cite{Wang2005}, that showed that any continuous piecewise linear function of $n$ variables can be represented by a sum of hinges containing at most $n+1$ linear functions. In the case of $f : \mathbb{R}\rightarrow \mathbb{R}$, \cite{Arora2018} shows a constructive method to build a 1-hidden-layer network having a maximum of $p$ units, where $p$ is the number of linear regions. However, this result does not extend to $n-$dimensional inputs. One of the contributions of our paper is to show a global linear approximation for piece-wise linear function $f : \mathbb{R}^{n_0}\rightarrow \mathbb{R}^{m}$ using 2-hidden-layer network that uses a fixed number of activation units that depend on the actual number of linear regions. 
     
    A necessary criterion for equivalent transformation is that the resulting network is as expressive as the original one. Methods to study neural network expressiveness include universal approximation theory~\cite{Cybenko1989}, VC dimension~\cite{Bartlett1998}, trajectory length~\cite{Raghu2017}, and linear regions~\cite{Serra2018,Montufar2014,Montufar2017}.
  
    More generally, our study of network approximation also relates to neural network compression. Neural network compression has been done through techniques like low-rank decomposition \cite{denton2014exploiting,jaderberg2014speeding,zhang2015efficient}, quantization \cite{rastegari2016xnor, courbariaux2016binarized, wu2016quantized}, architecture design~\cite{Szegedy2015, iandola2016squeezenet, howard2017mobilenets,huang2017densely, tang2018quantized}, pruning \cite{han2015learning,li2016pruning, molchanov2016pruning, han2016dsd}, sparse learning~\cite{liu2015sparse,zhou2016less,alvarez2016learning,wen2016learning}, and automatic discarding of layers in ResNets~\cite{Veit2017,yu2018learning,Herrmann2018}. 
\section{Notations and Preliminaries}
    We consider a \dnn with $L$ hidden layers, $n_0$ dimensional input vector $\vx=\{x_1,x_2,\dots,x_{n_0}\}$, and $n_{L+1}$  (i.e., $m =n_{L+1}$) dimensional output vector $\vy=\{y_1,\dots,y_{n_{L+1}}\}$. Each hidden layer $l$, $l \in \{1,2,\dots,L\}$, has $n_l$ hidden units with outputs given by $\{h_1^l,h_2^l,\dots,h_{n_l}^l\}$. We use $\text{ReLU}(p) = \max(0,p)$ as our activation function. The output of the hidden units and the output $\vy$ are defined as follows:
    \begin{align}
    h_i^{1}   &= \max\left(0,w^1_{i1} x_1 + \dots + w^1_{in_0} x_{n_0} + b^1_i\right), & \forall~i \in \{1,\dots,n_1\} \label{eq:formulation_start}\\
    h_i^{l+1} &= \max\left(0,w^{l+1}_{i1} h^{l}_1 + \dots + w^{l+1}_{in_l} h^{l}_{n_l} + b^{l+1}_i\right), & \forall~l \in \{1, \dots, L\}, i \in \{1,\dots,n_{l}\} \\
    y_i       &= \max\left(0,w^{L+1}_{i1} h^{l}_1 + \dots + w^{L+1}_{in_L} h^{l}_{n_L} + b^{L+1}_i\right), & \forall~i \in \{1,\dots,n_{L+1}\}\label{eq:formulation_end}
    \end{align}
    The terms $w^{l}_{ij}$ and $b^l_i$ denote the elements of the weight matrix $W^l$ and bias vector $b^l$, respectively. Each hidden neuron $i$ in layer $l+1$ is considered to be \emph{active} when $h_i^{l+1} = w^{l+1}_{11} h^{l}_1 + \dots w^{l+1}_{1n_l} h^{l}_{n_l} + b^{l+1}_i \ge 0$ and \emph{inactive} otherwise. We refer to a unit as \emph{stable} with respect to the input domain $D$ if the unit is always active or inactive for any $\vx \in D$, in which case we call it respectively as \emph{stably active} or \emph{stably inactive}. Otherwise, we denote the unit as \emph{unstable}.  We can identify the stable units by solving Mixed-Integer Linear Programming~(MILP) formulations for each unit~\cite{Cheng2017,FischettiMIP,tjeng2017evaluating,anderson2019strong}, which compute the maximum value $H^l_i$ of $W^l_i h^{l-1} + b_i^l$ and the maximum value $\bar{H}^l_i$ of $-\left(W^l_i h^{l-1} + b_i^l\right)$ for any input $\vx \in D$ (See Appendix~\ref{app.MILPforStableUnits}).
    
    Geometrically, each hidden unit $i$ in layer $l$ partitions its input space $h^{l-1}$ into two half-spaces divided by the \emph{activation hyperplane} $W^l_i h^{l-1} + b_i^l = 0$. Unit $i$ is active and produces a non-negative output in just one of those half-spaces. When all activation hyperplanes of layer $l$ are combined, they partition $h^{l-1}$ into \emph{possibly} an exponential number of what we denote \emph{linear regions}, each of which associated with a particular set of hidden units in the \dnn being active. We denote those sets of units as an \emph{activation set}. A linear region can be further partitioned by the units of subsequent layers, and consequently we associate each linear region of the network with a vector of activation sets for each layer, which we call an \emph{activation pattern}. Each linear region is a part of the input space that activates the same units in a network with piecewise-linear activations, such as ReLU, and thus the network defines to a piecewise-linear function with as many pieces as the number of linear regions.
    
    \textbf{Decision tree interpretation of the activation sets}
        Given an input $\vx$ we can denote the activations of various units in the first layer with the activation set $S_1$. Let us consider a small network with only 2 hidden units in each layer.
        \begin{wrapfigure}{r}{0.6\textwidth}
            \centering
            \begin{tikzpicture}[scale=0.65, every node/.style={scale=0.65}, every edge/.style={scale=0.65}]
\tikzset{vertex/.style = {shape=circle, draw=black!70, line width=0.1em, minimum size=1.4em}}
\tikzset{edge/.style = {->,> = latex'}}

	\node[] (t0) at (-1.2,10) {$\vx$};
	\node[vertex, fill=black!20] (h000)  at (-0.25,10)  {};

	\node[] (t1) at (-1.2,9) {$\vh^{1}$};
	\node[vertex, fill=my_green!55] (h010)  at (-0.5,9)  {};
	\node[vertex, fill=red!55] (h011)  at (0.0,9)  {};

	\node[] (t2) at (-1.2,8) {$\vh^{2}$};
	\node[vertex, fill=red!55] (h020)  at (-0.5,8)  {};
	\node[vertex, fill=my_green!55] (h021)  at (0.0,8)  {};

	\node[] (t3) at (-1.2,7) {$\vh^{3}$};
	\node[vertex, fill=red!55] (h030)  at (-0.5,7)  {};
	\node[vertex, fill=red!55] (h031)  at (0.0,7)  {};


	\path[black!30, draw, very thick, shorten <=2pt, shorten >=2pt]    (h000.south)  -- (h010.north);
	\path[black!30, draw, very thick, shorten <=2pt, shorten >=2pt]    (h000.south)  -- (h011.north);

	\path[black!30, draw, very thick, shorten <=2pt, shorten >=2pt]    (h010.south)  -- (h020.north);
	\path[black!30, draw, very thick, shorten <=2pt, shorten >=2pt]    (h011.south)  -- (h020.north);
	\path[black!30, draw, very thick, shorten <=2pt, shorten >=2pt]    (h010.south)  -- (h021.north);
	\path[black!30, draw, very thick, shorten <=2pt, shorten >=2pt]    (h011.south)  -- (h021.north);

	\path[black!30, draw, very thick, shorten <=2pt, shorten >=2pt]    (h020.south)  -- (h030.north);
	\path[black!30, draw, very thick, shorten <=2pt, shorten >=2pt]    (h021.south)  -- (h030.north);
	\path[black!30, draw, very thick, shorten <=2pt, shorten >=2pt]    (h020.south)  -- (h031.north);
	\path[black!30, draw, very thick, shorten <=2pt, shorten >=2pt]    (h021.south)  -- (h031.north);


	\node[vertex, fill=black!20] (h100)  at (6.0,10)  {$\vx$};

	\node[vertex, fill=white!55] (h110)  at (3.0,9)  {};
	\node[vertex, fill=cyan!75] (h111)  at (5.0,9)  {};
	\node[vertex, fill=white!55] (h112)  at (7.0,9)  {};
	\node[vertex, fill=white!55] (h113)  at (9.0,9)  {};

	\node[vertex, fill=white!55] (h120)  at (2.25,8)  {};
	\node[vertex, fill=white!55] (h121)  at (2.75,8)  {};
	\node[vertex, fill=white!55] (h122)  at (3.25,8)  {};
	\node[vertex, fill=white!55] (h123)  at (3.75,8)  {};
	\node[vertex, fill=white!55] (h124)  at (4.25,8)  {};
	\node[vertex, fill=white!55] (h125)  at (4.75,8)  {};
	\node[vertex, fill=cyan!75] (h126)  at (5.25,8)  {};
	\node[vertex, fill=white!55] (h127)  at (5.75,8)  {};
	\node[vertex, fill=white!55] (h128)  at (6.25,8)  {};
	\node[vertex, fill=white!55] (h129)  at (6.75,8)  {};
	\node[vertex, fill=white!55] (h1210)  at (7.25,8)  {};
	\node[vertex, fill=white!55] (h1211)  at (7.75,8)  {};
	\node[vertex, fill=white!55] (h1212)  at (8.25,8)  {};
	\node[vertex, fill=white!55] (h1213)  at (8.75,8)  {};
	\node[vertex, fill=white!55] (h1214)  at (9.25,8)  {};
	\node[vertex, fill=white!55] (h1215)  at (9.75,8)  {};

	\node[vertex, fill=black!70, scale=0.1] (d1351)  at (4.35,7)  {};
	\node[vertex, fill=black!70, scale=0.1] (d1352)  at (4.27,7)  {};
	\node[vertex, fill=black!70, scale=0.1] (d1353)  at (4.1899999999999995,7)  {};
	\node[vertex, fill=cyan!75] (h135)  at (4.75,7)  {};
	\node[vertex, fill=white!55] (h136)  at (5.25,7)  {};
	\node[vertex, fill=white!55] (h137)  at (5.75,7)  {};
	\node[vertex, fill=black!70, scale=0.1] (d1381)  at (6.65,7)  {};
	\node[vertex, fill=black!70, scale=0.1] (d1382)  at (6.73,7)  {};
	\node[vertex, fill=black!70, scale=0.1] (d1383)  at (6.8100000000000005,7)  {};
	\node[vertex, fill=white!55] (h138)  at (6.25,7)  {};

	\path[black!30, draw, very thick, shorten <=2pt, shorten >=2pt]    (h110.north)  -- node [midway, anchor=center, below, yshift=-0.05cm, xshift=-0.75cm] {\color{black!30}\tiny $\mathbf{(\emptyset)}$} (h100.south);
	\path[cyan!75, draw, ultra thick, shorten <=2pt, shorten >=2pt]    (h111.north)  -- node [midway, anchor=center, below, yshift=0cm, xshift=0.1cm] {\color{cyan!95}\tiny $\mathbf{(\{1\})}$} (h100.south);
	\path[black!30, draw, very thick, shorten <=2pt, shorten >=2pt]    (h112.north)  -- node [midway, anchor=center, below, yshift=0cm, xshift=0cm] {\color{black!30}\tiny $\mathbf{(\{2\})}$} (h100.south);
	\path[black!30, draw, very thick, shorten <=2pt, shorten >=2pt]    (h113.north)  -- node [midway, anchor=center, below, yshift=-0.0cm, xshift=2.2cm] {\color{black!30}\tiny $\mathbf{(\{1,2\})}$} (h100.south);

	\path[black!30, draw, very thick, shorten <=2pt, shorten >=2pt]    (h120.north)  -- node [midway, anchor=center, below, yshift=0.1cm, xshift=0cm] {\color{black!30}\tiny $\mathbf{~}$} (h110.south);
	\path[black!30, draw, very thick, shorten <=2pt, shorten >=2pt]    (h121.north)  -- node [midway, anchor=center, below, yshift=0.1cm, xshift=0cm] {\color{black!30}\tiny $\mathbf{~}$} (h110.south);
	\path[black!30, draw, very thick, shorten <=2pt, shorten >=2pt]    (h122.north)  -- node [midway, anchor=center, below, yshift=0.1cm, xshift=0cm] {\color{black!30}\tiny $\mathbf{~}$} (h110.south);
	\path[black!30, draw, very thick, shorten <=2pt, shorten >=2pt]    (h123.north)  -- node [midway, anchor=center, below, yshift=0.1cm, xshift=0cm] {\color{black!30}\tiny $\mathbf{~}$} (h110.south);
	\path[black!30, draw, very thick, shorten <=2pt, shorten >=2pt]    (h124.north)  -- node [midway, anchor=center, below, yshift=0.1cm, xshift=0cm] {\color{black!30}\tiny $\mathbf{~}$} (h111.south);
	\path[black!30, draw, very thick, shorten <=2pt, shorten >=2pt]    (h125.north)  -- node [midway, anchor=center, below, yshift=0.1cm, xshift=0cm] {\color{black!30}\tiny $\mathbf{~}$} (h111.south);
	\path[black!20, draw, very thick, shorten <=2pt, shorten >=2pt]    (h127.north)  -- node [midway, anchor=center, below, yshift=0.1cm, xshift=0cm] {\color{black!30}\tiny $\mathbf{~}$} (h111.south);
	\path[cyan!75, draw, ultra thick, shorten <=2pt, shorten >=2pt]    (h126.north)  -- node [midway, anchor=center, below, yshift=0.15cm, xshift=0.7cm] {\color{cyan!95}\tiny $\mathbf{(\{1\},\{2\})}$} (h111.south);
	
	\path[black!30, draw, very thick, shorten <=2pt, shorten >=2pt]    (h128.north)  -- node [midway, anchor=center, below, yshift=0.1cm, xshift=0cm] {\color{black!30}\tiny $\mathbf{~}$} (h112.south);
	\path[black!30, draw, very thick, shorten <=2pt, shorten >=2pt]    (h129.north)  -- node [midway, anchor=center, below, yshift=0.1cm, xshift=0cm] {\color{black!30}\tiny $\mathbf{~}$} (h112.south);
	\path[black!30, draw, very thick, shorten <=2pt, shorten >=2pt]    (h1210.north)  -- node [midway, anchor=center, below, yshift=0.1cm, xshift=0cm] {\color{black!30}\tiny $\mathbf{~}$} (h112.south);
	\path[black!30, draw, very thick, shorten <=2pt, shorten >=2pt]    (h1211.north)  -- node [midway, anchor=center, below, yshift=0.1cm, xshift=0cm] {\color{black!30}\tiny $\mathbf{~}$} (h112.south);
	\path[black!30, draw, very thick, shorten <=2pt, shorten >=2pt]    (h1212.north)  -- node [midway, anchor=center, below, yshift=0.1cm, xshift=0cm] {\color{black!30}\tiny $\mathbf{~}$} (h113.south);
	\path[black!30, draw, very thick, shorten <=2pt, shorten >=2pt]    (h1213.north)  -- node [midway, anchor=center, below, yshift=0.1cm, xshift=0cm] {\color{black!30}\tiny $\mathbf{~}$} (h113.south);
	\path[black!30, draw, very thick, shorten <=2pt, shorten >=2pt]    (h1214.north)  -- node [midway, anchor=center, below, yshift=0.1cm, xshift=0cm] {\color{black!30}\tiny $\mathbf{~}$} (h113.south);
	\path[black!30, draw, very thick, shorten <=2pt, shorten >=2pt]    (h1215.north)  -- node [midway, anchor=center, below, yshift=0.1cm, xshift=0cm] {\color{black!30}\tiny $\mathbf{~}$} (h113.south);

	\path[cyan!75, draw, ultra thick, shorten <=2pt, shorten >=2pt]    (h135.north)  -- node [midway, anchor=center, below, yshift=0.1cm, xshift=-0.99cm] {\color{cyan!95}\tiny $\mathbf{(\{1\},\{2\},\emptyset)}$} (h126.south);
	\path[black!30, draw, very thick, shorten <=2pt, shorten >=2pt]    (h136.north)  -- node [midway, anchor=center, below, yshift=0.1cm, xshift=0cm] {\color{black!30}\tiny $\mathbf{~}$} (h126.south);
	\path[black!30, draw, very thick, shorten <=2pt, shorten >=2pt]    (h137.north)  -- node [midway, anchor=center, below, yshift=0.1cm, xshift=0cm] {\color{black!30}\tiny $\mathbf{~}$} (h126.south);
	\path[black!30, draw, very thick, shorten <=2pt, shorten >=2pt]    (h138.north)  -- node [midway, anchor=center, below, yshift=0.1cm, xshift=0cm] {\color{black!30}\tiny $\mathbf{~}$} (h126.south);
	\node at (1,8.5) [left color=blue!50!cyan!95, right color=blue!50!cyan!95,single arrow, minimum height=1cm ] {};
\end{tikzpicture}
            \caption{We show the decision tree interpretation of one of the activation pattern $(\{1\}, \{2\}, \emptyset)$. The green and red colors denote the active and inactive states of the units, respectively.}
            \label{fg.decision_tree}
            \vspace{-0.35cm}
        \end{wrapfigure}
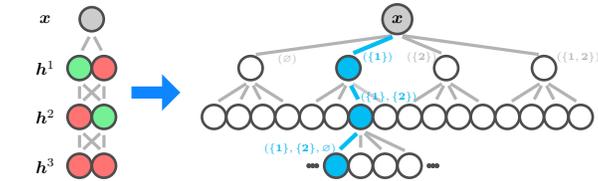 The activation sets $S_1$, $S_2$, and $S_3$ can be one of the following: $\emptyset$, $\{1\}$, $\{2\}$, and $\{1,2\}$. It is easy to see the activation patterns (concatenation of activation sets from the different layers) in the form of a decision tree as shown in Figure~\ref{fg.decision_tree}.
        Given an input $\vx$ we have four possible outputs from the first hidden layer - $S_1=\emptyset:h^1_1=0,h^1_2=0; S_1=\{1\}:h^1_1>0,h^1_2=0; S_1=\{2\}:h^1_1=0,h^1_2>0; S_1=\{1,2\}:h^1_1>0,h^1_2>0$.
        From the second hidden layer, we can have 16 possible activation patterns from the concatenation of 4 possible values for $S_1$ and 4 possible values for $S_2$ as shown in Figure~\ref{fg.decision_tree}. However, in reality there are significantly smaller number of feasible activation patterns as given by Zaslavsky's Theorem~\cite{Zaslavsky1975} and other results in linear regions~\cite{Serra2018,Raghu2017,Montufar2014,Montufar2017} (See Appendix~\ref{app.linear_regions}).
\section{Exact Global and Local Transformations}
    Algorithm~\ref{alg:compress}, which we denote \texttt{StabilityCompression}, 
    removes stable units and layers while adjusting the weights and biases of other units to keep the original and the resulting neural networks equivalent on a given domain $D$. The algorithm loops over the layers to check which units are stable. First, stably inactive units are removed as long as there are other units left in the layer. Similarly, stably active units are removed if their weights are linearly dependent on the weights of stably active units that have been previously inspected in the same layer. If all remaining units in a layer are stably active, then the layer is removed by directly joining the layers before and after it. In the particular case in which only one stably inactive unit is left in a layer, then all hidden layers are removed. 
    
    \setlength{\textfloatsep}{0pt} 
    \begin{algorithm}[!htb]
    \caption{Based on the stability of the units for a input domain $D$, removes units and layers while keeping the resulting neural network locally equivalent with respect to $D$}
    \label{alg:compress}
    {\AlgorithmFontSize
    \begin{algorithmic}[1]
    \Procedure{StabilityCompression}{$D$}
    \For{$l \gets 1, \ldots, L$}
    \State $A \gets \{ \}$ \Comment{Set of stably active units in layer $l$}
    \State Unstable $\gets$ False \Comment{If there are unstable units in layer $l$}
    \For{$i \gets 1, \ldots, n_l$}
    \State Compute $H_i^l$ and $\bar{H}_i^l$ with respect to $D$
    \If{$H_i^l \leq 0$} \label{lin:inactive}
    \Comment{Unit $i$ is stably inactive}
    \If{$i < n_l$ \textbf{or} $|A| > 0$ \textbf{or} Unstable} \label{lin:one_inactive}
    \Comment{Layer $l$ still has other units} 
    \State Remove unit $i$ from layer $l$
    \Comment{Remove unit $i$}
    \EndIf
    \ElsIf{$\bar{H}_i^l \leq 0$} \label{lin:active}
    \Comment{Unit $i$ is stably active}
    \If{rank$\left(W^l_{A \cup \{ i \}}\right)$ > $|A|$} \Comment{$w^l_i$ is linearly independent from rows in $W^l_A$} \label{lin:li_active}
    \State $A \gets A \cup \{ i \}$ \Comment{Keep unit in the network}
    \ElsIf{$i = n_l$ \textbf{and} $|A|=0$ \textbf{and not} Unstable} \Comment{Last unit in layer $l$ has $w^l_i =$ \textbf{0}}
    \State $A \gets A \cup \{ i \}$ \Comment{Keep unit until block from line~\ref{lin:all_stable}}
    \Else \Comment{Output of unit $i$ is linearly dependent} \label{lin:ld_active}
    \State Find $\{ \alpha_k \}_{k \in A}$ such that $w^l_i = \sum_{k \in A} \alpha_k w^l_k$
    \For{$j \gets 1, \ldots, n_{l+1}$} \Comment{Adjust activation functions with units in $A$}
    %
    \State $w^{l+1}_{jk} \gets w^{l+1}_{jk} + \sum_{k \in A} \alpha_k w^{l+1}_{ji}$
    \State $b^{l+1}_j \gets b^{l+1}_j + w^{l+1}_{ji} (b^l_i + \sum_{k \in A} \alpha_k  b^l_k)$
    \EndFor
    \State Remove unit $i$ from layer $l$ 
    \Comment{Unit $i$ is no longer necessary}
    \EndIf
    \Else \label{lin:unstable}
    \State Unstable $\gets$ True 
    \Comment{At least one unit is not stable}
    \EndIf
    \EndFor
    \If{\textbf{not} Unstable} \Comment{All units left in layer $l$ are stable}
    \If{$|A|>0$}  \label{lin:all_stable}
    \Comment{All units left in layer $l$ are stably active}
    \State Create matrix $\bar{W} \in \mathbb{R}^{n_{l-1} \times n_{l+1}}$ and vector $\bar{b} \in \mathbb{R}^{n_{l+1}}$
    \For{$i \gets 1, \ldots, n_{l+1}$} \Comment{Compute parameters to directly connect layers $l-1$ and $l+1$}
    %
    \State $\bar{b}_i \gets b^{l+1}_i + \sum_{k \in A} w^{l+1}_{ik} b^l_k$
    \For{$j \gets 1, \ldots, n_{l-1}$}
    \State $\bar{w}_{ij} \gets \sum_{k \in A} w^l_{kj} w^{l+1}_{ik}$
    \EndFor
    \EndFor
    \State Remove layer $l$, replace weights and biases in next layer with $\bar{W}$ and $\bar{b}$ 
    \Else \label{lin:last_inacive}
    \Comment{Last unit in layer $l$ is stably inactive}
    \State Compute output $\Upsilon$ for any input $\chi \in D$ \Comment{Network function is constant in $D$}
    \State Remove layers 1 to $L$ \Comment{Remove all hidden layers}
    \For{$j \gets 1, \ldots, n_{L+1}$} \Comment{Set constant values in output layer}
    \State $w^{L+1}_j \gets$ \textbf{0}
    \State $b^{L+1}_j \gets \Upsilon_j$
    \EndFor
    \State \textbf{return} \Comment{Leave early}
    \EndIf
    \EndIf
    \EndFor
    \EndProcedure
    \end{algorithmic}
    }
    \end{algorithm}

    \begin{theorem}\label{th:1}
    For a given neural network \dnnone, Algorithm~\ref{alg:compress} produces a neural network \dnntwo such that \dnnone and \dnntwo are local equivalent with respect to the input domain $D$. 
    \end{theorem}
    
    Please refer to Appendix~\ref{app.proof:1} for the detailed proof.

\section{Global Linear Approximation to a 2-Hidden-Layer Network}
\begin{figure}[!htb]
        \begin{center}
            \input{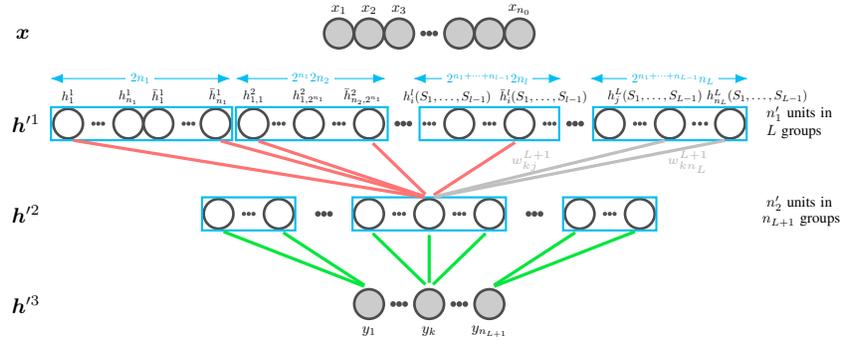}
        \end{center}
        \caption{\emph{Global linear transformation of an L-hidden-layer network to a 2-hidden-layer shallow network}. We show the general form of the shallow network where the first layer consists of all possible hyperplanes by considering every hidden unit to either be active or inactive. The second layer computes the hidden units in the $(L-1)$th layer for all possible activation patterns. The output is the disjunction of the outputs for all possible activation patterns. The weights are designed in such a manner that all the outputs from the linear regions, not containing the input point $\mathbf{x}$, are forced to zeros. }
        \label{fg.general_case}
    \end{figure}

    We show an approximate global transformation in the following theorem, where the original ReLU network is approximated by a ReLU network with only 2 hidden layers.
    \begin{theorem}\label{th:2}
        Given a \dnnone~with associated function $\fone:\mathbb{R}^{n_0}\rightarrow \mathbb{R}^{n_{L+1}}$ with $L$ hidden layers and $n_l$ units in layer $l$ (see Figure~\ref{fg.intro_figure}), there exists a global linear approximation \dnntwo~with associated function $\ftwo:\mathbb{R}^{n_0}\rightarrow \mathbb{R}^{n_{L+1}}$ such that, for a given norm $\ell$ and distance $\epsilon$, $\fone(\vx)=\ftwo(\vx)$ for any $\vx \in \mathbb{R}^{n_0}$ such that $\fone$ is a linear function on $\{ \vx' : \|\vx-\vx'\|_{\ell} \le \epsilon\}$. The widths of the transformed network \dnntwo are shown in Figure~\ref{fg.general_case}:
        \begin{align}
            n'_1 &=  \sum\limits_{l=1}^{L-1} \left(2^{n_1+\dots+n_{l-1}}\right)2n_l + \left(2^{n_1+\dots+n_{L-1}}\right)n_L \label{eq.NoUnits1stLayer} \\ \text{~and~~~~}
            n'_2 &= \left(2^{n_1+\dots+n_{L-1}}\right)n_{L+1} \label{eq.NoUnits2ndLayer}
        \end{align}
        \label{th.general_case}
    \end{theorem}
    We prove this theorem by constructing a 2-hidden-layer network \dnntwo with associated function $\ftwo:\mathbb{R}^{n_0}\rightarrow \mathbb{R}^{n_{L+1}}$ that is a global linear approximation of the original $L$-hidden-layer \dnnone with associated function $\fone:\mathbb{R}^{n_0}\rightarrow \mathbb{R}^{n_{L+1}}$ for a given maximum offset parameter $\epsilon$. The theorem essentially says that for every point $\vx$ that is at least $\epsilon$ distance from the boundary of the linear regions, we have $\fone(\vx) = \ftwo(\vx)$. Please refer to Appendix~\ref{app.proof:2} for the detailed proof. The comparison of the number of units, non-zero weights as well as non-zero biases in the original and the transformed ReLU networks is carried out in Appendix~\ref{app.analysis_net_th2}.

    \subsection{Transformation using Activation Patterns}
        Theorem~\ref{th:2} considers the possibility of all $2^k$ activation patterns where $k$ is total number of hidden units in the network. However, the number of feasible activation patterns is generally quite less and has been studied in the literature~\cite{Serra2018,Raghu2017,Montufar2014,Montufar2017}. 
        Given the set of all feasible activation patterns ${\cal A}$, we show a more compact 2-hidden-layer global linear approximation to the original network in the following theorem.
        \begin{theorem}\label{th:3}
            Given a \dnnone~with associated function $\fone:\mathbb{R}^{n_0}\rightarrow \mathbb{R}^{n_{L+1}}$ with $L$ hidden layers, $n_l$ units in each layer $l$, and a set of feasible activation patterns ${\cal A}$, there exists a network \dnntwo~with associated function $\ftwo:\mathbb{R}^{n_0}\rightarrow \mathbb{R}^{n_{L+1}}$   
            that is global linear approximation of \dnnone. 
            The transformed network \dnntwo has 2 hidden layers 
            with widths $(n'_1,n'_2)$ given by
            \begin{align}
                n'_1 &= \sum_{l=1}^{L-1} |{\cal A}_{l-1}|2n_l  + |{\cal A}_{L-1}|n_L \label{eq.NoUnits1stLayer_ap}\\ \text{~and~~~~}
                n'_2 &= |{\cal A}_{L-1}|n_{L+1} \label{eq.NoUnits2ndLayer_ap}
            \end{align}
            Set ${\cal A}_l$ consist of the feasible activation patterns up to layer $l$, and any such subset $B \in {\cal A}_l$ is then given by 
            $    
            B = A - \underbrace{\{\emptyset,\dots,\emptyset,{\cal P}^{l+1},\dots,{\cal P}^{L}\}}_{L~sets},\forall~A \in {\cal A}
            $, where
            ${\cal P}^l$ denotes the power set of $\{1,2,...,n_l\}$.
            \label{eq.general_case_with_ap}
        \end{theorem}
        
        Similar to the previous theorem, we prove this result by constructing a 2-hidden layer network \dnntwo with associated function $\ftwo:\mathbb{R}^{n_0}\rightarrow \mathbb{R}^{n_{L+1}}$ that is a global linear approximation of the original $L$-layer \dnnone with associated function $\fone:\mathbb{R}^{n_0}\rightarrow \mathbb{R}^{n_{L+1}}$ for a given maximum offset parameter $\epsilon$. The transformed 2-layer network is similar to the one constructed by using Theorem~\ref{th:2} and shown in Figure~\ref{fg.general_case}, except that we discard all the hidden units in the transformed network that are unnecessary. In other words, the set of feasible activation patterns ${\cal A}$ allows us to discard all the units that correspond to infeasible activations. Please refer to Appendix~\ref{app.proof:3} for the proof.
\section{Experiments}
    We conducted proof-of-concept experiments using the MNIST dataset.
    We use \dnn architectures with input size 784 and 10 output units. These networks have varying number of hidden layers and hidden units per layer. The exact numbers vary per experiment performed. All our models use ReLU activation functions and softmax as the loss function. Unless stated otherwise, we use $l_1$ regularization with a weight of $0.003$ on the first layer and $0$ for the remaining layers as in \cite{xiao2018training}. The weight decay is kept at $0$ unless otherwise stated. For models trained with adversarial examples, PGD attacks are carried out with \lInfAttackD{0.15}{200}{0.1} as in \cite{xiao2018training}. We use a batch size of $64$ and SGD with a learning rate of $0.01$ and momentum of $0.9$ for training the model to $120$ epochs. The learning rate is decayed by a factor of $0.1$ after every $50$ epochs. The weights of the network are initialized with the Kaiming initialization \cite{He2016DeepRL} and the biases are initialized to zero. The models are trained in Pytorch~\cite{paszke2017automatic} and Gurobi~\cite{gurobi} is used as the MILP solver in order to identify which units are stable and to enumerate linear regions when necessary. 

    \subsection{Equivalent Networks from Applying Algorithm~\ref{alg:compress}} 
    
    Our first experiment illustrates the potential to prune units in neural network by identifying those that are stable. We use a network with two hidden layers, each containing from 50 to 500 units, which is denoted as \advnet-\mlp{2}{n} in \cite{xiao2018training}. 
    In preliminary evaluations, we confirmed the finding in~\cite{xiao2018training} that  networks trained with adversarial examples generated by PGD and $l_1$ regularization lead to more stable units. Please refer to Appendix~\ref{app.act_with_regularization} for more experiments on different types of regularization.
        
\begin{figure}[!htb]
        \centering
        \includegraphics[width=0.65\linewidth]{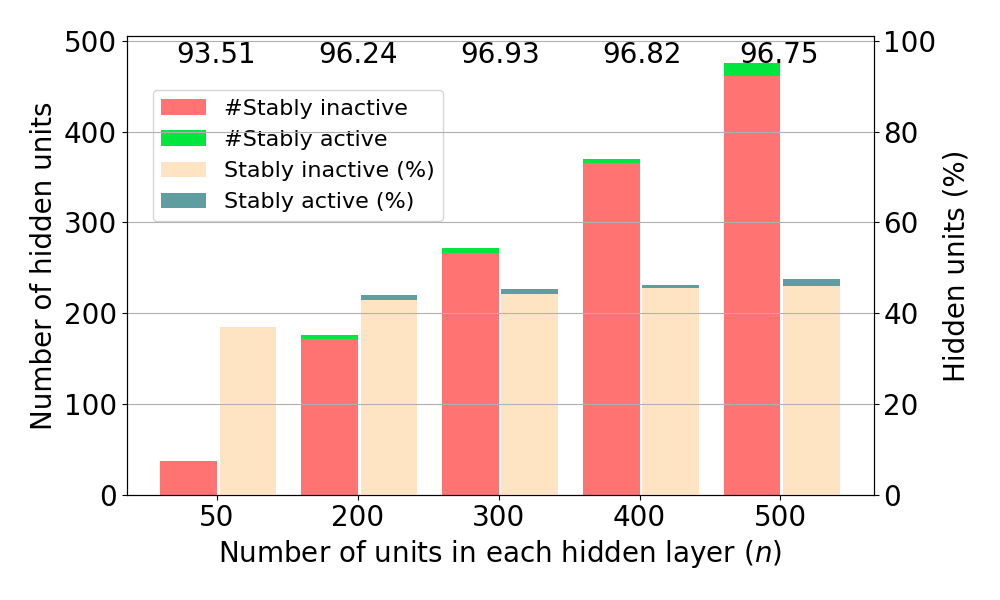}
        \caption{Number as well as percentage of stable (inactive and active) hidden units in \advnet-\mlp{2}{n} networks according to layer width. The validation accuracies for each $n$ is shown on the top.}
        \label{fig:inactive_with_w}
    \end{figure}

    Figure \ref{fig:inactive_with_w} shows the number of stably inactive units and their percentage with respect to the  total number of hidden units in \advnet-\mlp{2}{n} networks with the number of units in each hidden layer (width $n$). The percentage of stably inactive units as a percentage of the total hidden units is more than $40\%$ and is fairly constant as we vary the width $n$. As shown in Algorithm~\ref{alg:compress}, we easily discard the stably inactive units and obtain smaller networks that are equivalent to the original one. 
    
    \subsection{Global Linear Approximation with Theorems~\ref{th:2} and \ref{th:3}} 
    
    Our second experiment uses small networks to show the result of the approximate transformations characterized by Theorems~\ref{th:2} and \ref{th:3}. In particular, we show the significant reduction in the size of the resulting network when the activation patterns associated with linear regions are leveraged. 
    
    We train networks with $l_1$ regularization with and without adversarial examples generated by PGD (denoted by Adv in Table~\ref{tab:transformation_activation}). Some of the networks with a small number of hidden layers and small number of depths in the hidden layer did not converge with adversarial training (see Appendix~\ref{app.act_with_regularization}). We then use Theorem~\ref{th:2} to transform these networks into 2-hidden-layer networks. 
    For the global linear approximation using Theorem~\ref{th:3}, we also identify the possible activation patterns associated with the linear regions of the network, which we compute using MILP~\cite{Serra2018} (see Appendix~\ref{app.MILPforlinRegions} for more details).
    
    Table~\ref{tab:transformation_activation} shows the number of hidden units in the first and second hidden layer of the transformed network with and without activation patterns. The models  trained with adversarial examples and $l_1$ regularization have very small number of activation patterns (see Appendix~\ref{app.act_with_regularization}). As shown in the table, the transformation that uses the actual linear regions can be significantly smaller compared to the one achieved without using the linear regions. 

        \begin{table}[!htb]
            \centering
            \caption{Number of units per layer of the global linear approximation from Theorems~\ref{th:2} and \ref{th:3} according to the the original network architecture, training method, and number of activation patterns. Note that there is a significant decrease in the number of activation units while using the linear regions.}
            \label{tab:transformation_activation}
            {\setlength{\tabcolsep}{0.2em}
            \begin{tabular}{|c| c| c!{\vrule width 1.5pt} c| c!{\vrule width 1.5pt} c| c!{\vrule width 1.5pt}} 
                \ChangeRT{0.8pt}
                
                \rule{0pt}{10pt}
                \multirow{2}{*}{\textbf{Architecture}} &
                \textbf{Training} &
                \textbf{Activation} &
                \multicolumn{2}{c!{\vrule width 1.5pt}}{\textbf{Theorem~\ref{th:2}}} &
                \multicolumn{2}{c!{\vrule width 1.5pt}}{~~\textbf{Theorem~\ref{th:3}}~} \\[0.5ex]\cline{4-7}
                
                & \textbf{Method} 
                & \textbf{Patterns} & $n'_1$ & $n'_2$ & $n'_1$ & $n'_2$\\[0.5ex]
                
                \ChangeRT{0.8pt}
                $[$784, 5, 5, 5, 10$]$ & $l_1$ & 458 & 5450 & 10240 & 1540 & 1900\\
                $[$784, 5, 5, 10, 10, 10$]$ & $l_1$ & 1696 & 10506570 & 10485760  & 5060 & 4050\\
                $[$784, 5, 5, 5, 10, 10, 10$]$ & $l_1$ + Adv & 354 & 336210250 & 335544320 & 2620 & 1320\\
                \ChangeRT{0.8pt} 
            \end{tabular}}
        \end{table}

    \subsection{Local Stability Analysis} 
        Our last experiment illustrates the potential to further prune a neural network when the input is restricted to a smaller domain. In this experiment, we choose a point $\vx = \bar{\vx}$ defined by an image in the validation set and we bound the set of valid inputs to the hypercube defined by $\{ x : |x_i - \bar{x}_i| \le \delta ~\forall i=1,\ldots,n_0\}$ for some positive constant $\delta$. The smaller $\delta$ is, the more units are stable and the fewer activation patterns correspond to feasible linear regions. Similarly, a global linear approximation also becomes considerably smaller in such case. 
        
\begin{figure}[!htb]
            \begin{subfigure}{.33\textwidth}
              \centering
              \includegraphics[width=\linewidth]{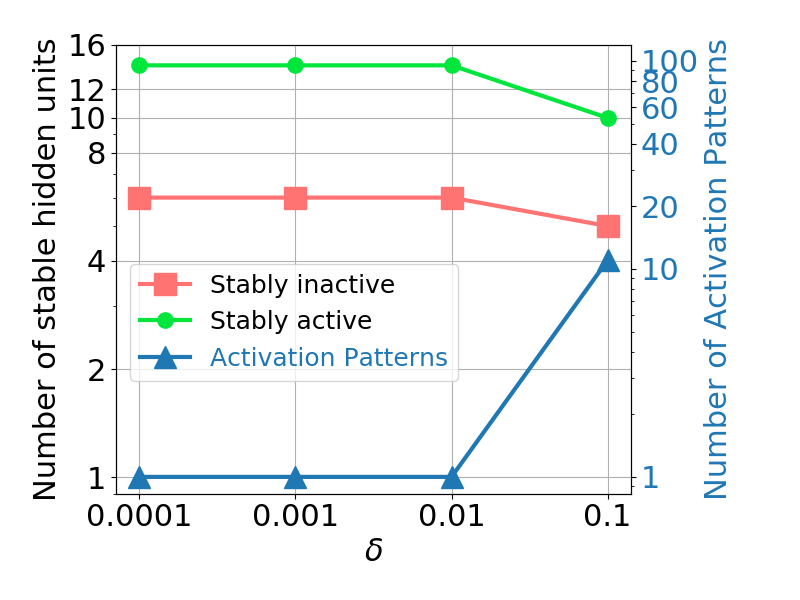}
              \caption{Class 3}
            \end{subfigure}%
            \begin{subfigure}{.33\textwidth}
              \centering
              \includegraphics[width=\linewidth]{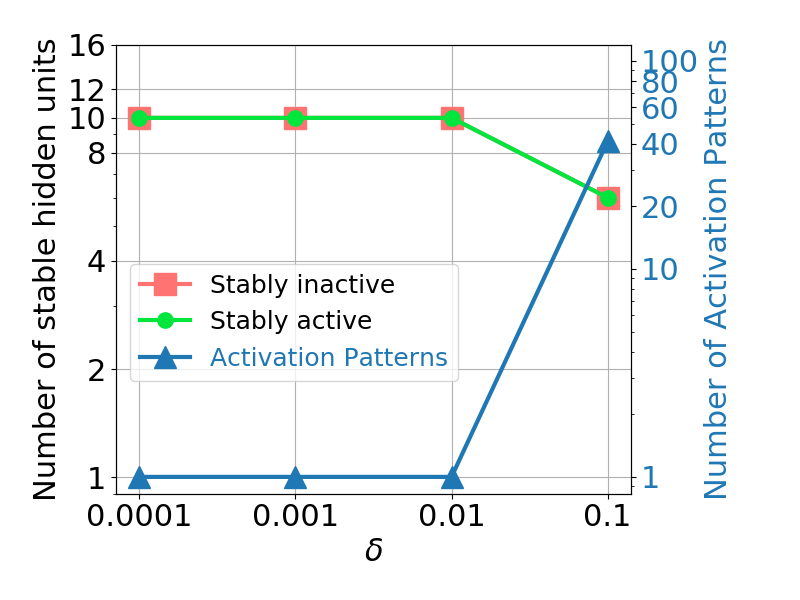}
              \caption{Class 6}
            \end{subfigure}%
            \begin{subfigure}{.33\textwidth}
              \centering
              \includegraphics[width=\linewidth]{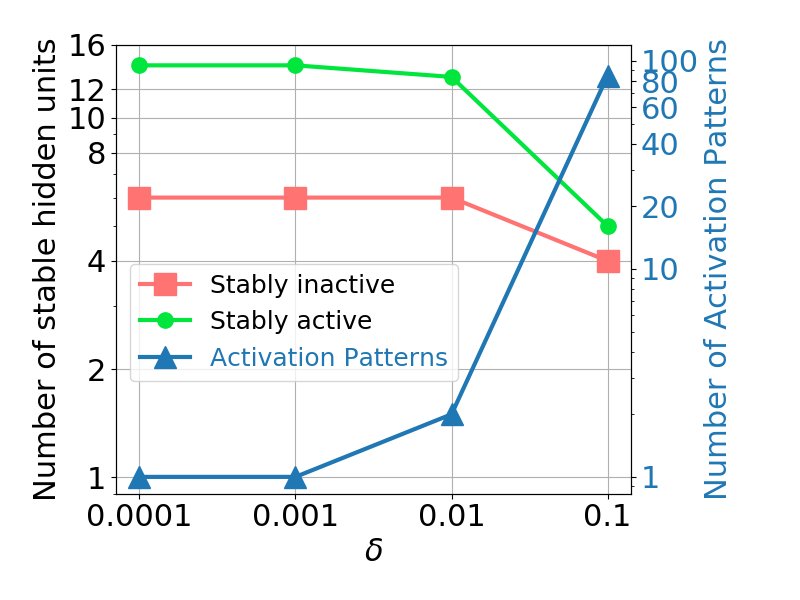}
              \caption{Class 9}
            \end{subfigure}
            \caption{Variation in the number of stably inactive and stably active hidden units as well as activation patterns with $\delta$ for a randomly sampled example from few of the classes. The y-axes are in log-scale. (See Appendix~\ref{app.local_stab} for all classes.)}
            \label{fig:loc_stab_vs_delta}
        \end{figure}

        We carry out the local stability analysis of the network with architecture $[$784, 5, 5, 5, 5, 10$]$ trained with $l_1$ regularization. We randomly choose one value of $\bar{\vx}$ for each class of the validation set. 
        Figure~\ref{fig:loc_stab_vs_delta} shows the number of stably inactive and stably active hidden units as well as the number of activation patterns according to $\delta$ for some of the MNIST classes. When $\delta$ increases,  the number of activation patterns increases and the number of stably inactive hidden units of the network decreases. 
        
        We also perform the same analysis by choosing $\bar{x}_i = \delta$ and evaluating the number of stable units and activation patterns corresponding to feasible linear regions as we vary the length $\alpha = 2 \delta$ of the hypercube along each coordinate. The results are presented in Table~\ref{tab:local_stab_analysis_2}.
        
        \begin{table}[!htb]
            \centering
            \caption{Number of stably inactive and active nodes as well as activation patterns based on the upper bound on the inputs.}
            \label{tab:local_stab_analysis_2}
            \begin{tabular}{|c| c| c| c| c|}
                \ChangeRT{0.8pt}
                $\mathbf{\alpha}$ & \textbf{Inactive} & \textbf{Active} & \textbf{Activation Patterns} & \textbf{Time} (s) \\ [0.5ex]
                \ChangeRT{0.8pt}
                0.0001 & 5 & 15 & 1   & 0.011 \\
                0.001  & 5 & 15 & 1   & 0.011\\
                0.01   & 4 & 15 & 2   & 0.015\\
                0.1    & 4 & 5  & 64  & 0.337\\
                1      & 3 & 0  & 458 & 1.066\\
                \ChangeRT{0.8pt}
            \end{tabular}
        \end{table}         
\section{Discussion}
    In this paper we have shown that the stability of ReLUs can be used for loss-less compression of neural networks. In other words, we can remove units and even layers while obtaining a resulting neural network that represents the same function as before. We are not aware of any prior work that computes and utilizes stable units for model compression. 
    
    During the experiments we observed that the number of linear regions decrease in many training settings: (a) $l_1$ regularization (b) using dropout with $l_1$ or $l_2$ regularization, (c) adversarial training with $l_1$ regularization. Furthermore, we observed the largest decrease in the number of linear regions while using adversarial training with $l_1$ regularization. The decrease in the number of linear regions due to  regularization validates the intuition that linear regions is typically considered to be a proxy for network expressiveness~\cite{Montufar2017,Raghu2017,Serra2018}. In a sense, we restrict the expressiveness of a network when we perform regularization.
    
    We experimentally observed that the global linear approximation using 2 layers is considerably more compact when using the linear regions. Arora \etal~\cite{Arora2018} has already shown a constructive method for functions $f : \mathbb{R}\rightarrow \mathbb{R}$ with one-dimensional input to build a one-hidden-layer network that depends on the number of linear regions. However, their results for $n-$dimensional input uses a deeper network with $\ceil{log(n_0+1)}+1$ layers without relying on the number of linear regions. On the contrary, we show a global linear transformation for $n-$ dimensional input using 2-hidden layers and the network size depends on the number of linear regions. These results indicate that the number of linear regions would play a critical role if we focus on building a shallow equivalent network. Despite many prior results showing the existence of equivalent shallow networks for threshold and sigmoid units, most of them do not provide any algorithm to perform the actual transformation, nor provide the actual analytical expression for the size of the final network. This is understandable due to the lack of techniques/methods to actually compute the possible linear regions even for small-scale networks, until recently~\cite{Serra2018}. 
    
    While this work demonstrates the general idea of equivalent networks and global approximations using linear regions for small-scale networks, our future work would focus on finding techniques to extend this for CNNs and large-scale networks.
    
    \bibliographystyle{unsrt}

    \begin{appendices}
        \newpage
\appendix

\textbf{APPENDIX}\\
In this supplementary document, we provide
\begin{itemize}
    \item Appendix \ref{app.linear_regions}: Definition of linear region and its relation to activation patterns in a ReLU \dnn.
    \item Appendix \ref{app.MILPforStableUnits}: MILP formulation for getting the maximum and minimum bounds of each hidden unit of a ReLU \dnn.
    \item Appendix \ref{app.MILPforlinRegions}: MILP formulation for enumerating all feasible activation patterns of a ReLU \dnn.
    \item Appendix \ref{app.proof:1}: Proof of Theorem~\ref{th:1}.
    \item Appendix \ref{app.proof:2}: Proof of Theorem~\ref{th:2}.
    \item Appendix \ref{app.analysis_net_th2}: Comparison of the number of units, non-zero weights as well as non-zero biases in the original and the transformed ReLU \dnn.
    \item Appendix \ref{app.proof:3}: Proof of Theorem~\ref{th:3}.
    \item Appendix \ref{app.act_with_regularization}: Effect of different regularization on the number of activation patterns of a network.
    \item Appendix \ref{app.local_stab}: Variation in the number of activation patterns and stably inactive hidden units with $\delta$ which can be used for local transformations.
\end{itemize}

\section{Linear Regions}\label{app.linear_regions}
\begin{figure}[!htbp]
        \centering
        \includegraphics[width=\linewidth]{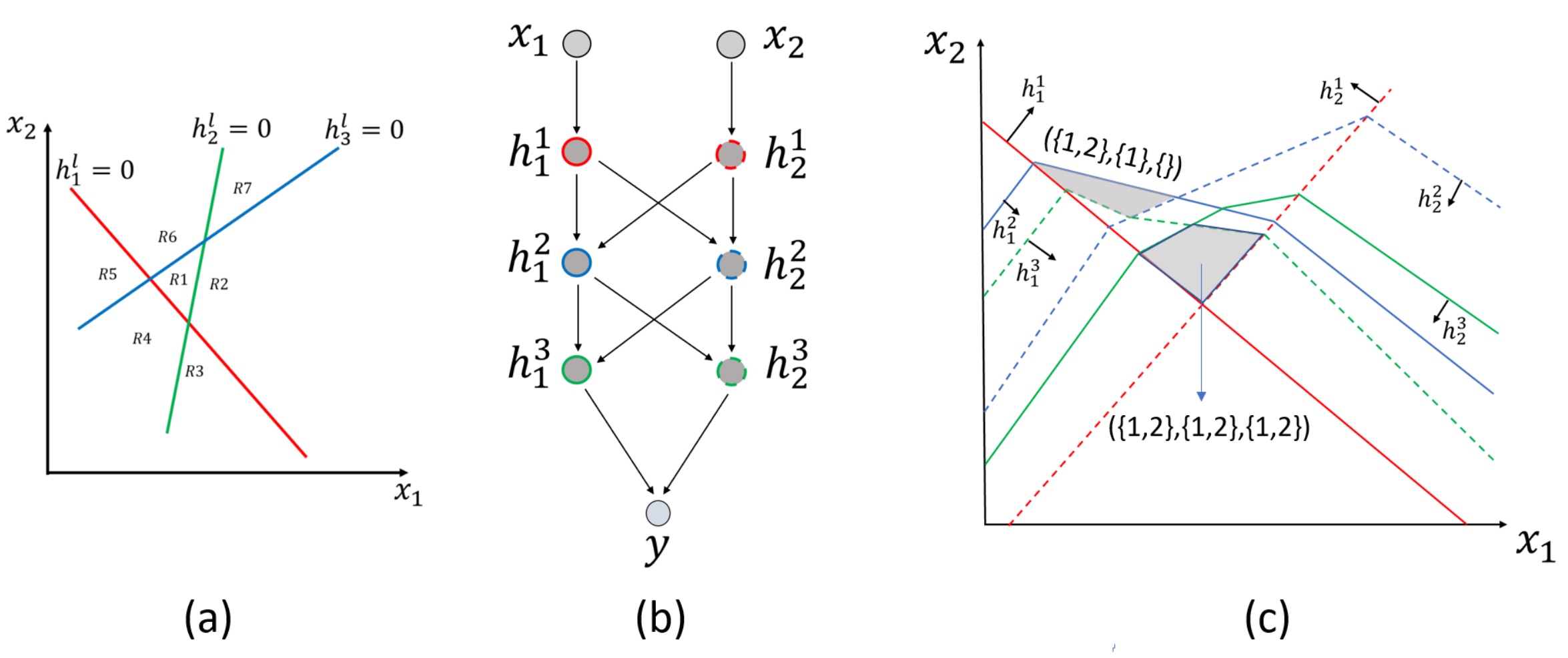}
        \caption{(a) Three hyperplanes dividing a 2D space into 7 regions as given by Zaslavsky's Theorem~\cite{Zaslavsky1975}. (b) A simple network with two inputs, three hidden layers, and one output. (c) Recursive subdivision of input space into linear regions by the three hidden layers shown on the 2D input space.}
        \label{fg.background_figure}
    \end{figure}

    \paragraph{Hyperplane Arrangements}
        Consider a set of 3 hyperplanes given by the equations $h^l_1$, $h^l_2$, and $h^l_3$ as shown in Figure~\ref{fg.background_figure}(a). The natural question is to understand all possible states of the associated hidden units. This question was answered almost half a century earlier by Zaslavsky~\cite{Zaslavsky1975}, and this number is always smaller than $2^n$ while using $n$ hyperplanes. Zaslavsky~\cite{Zaslavsky1975} shows that an arrangement of $n$ hyperplanes partitions a $d$-dimensional space into at most $\sum_{s=0}^d {n \choose s}$ regions. This bound is tight in general position, where a small perturbation of the hyperplanes do not change the number of partitions. An illustration of Zaslavsky's Theorem is shown in Figure~\ref{fg.background_figure}(a) where 3 hyperplanes produce 7 regions $\sum_{s=0}^2 {3 \choose s}={3 \choose 0} + {3 \choose 1} + {3 \choose 2} = 1 + 3 + 3 = 7$.
    
    \paragraph{Linear Regions and Activation Patterns}
        Given the input vector $\vx = \{x_1,\ldots,x_{n_0}\}$, for every layer $l$ we define an activation set $S^l \subseteq \{1,\ldots, n_l\}$ such that $e \in S^l$ if and only if the ReLU $e$ is active, i.e. $h^l_e > 0$. The aggregation of the activation sets for all the layers form the activation pattern $\mathcal{S} = (S^1, \ldots, S^{l})$. Let us denote the power set of $\{1,2,...,n_l\}$ by ${\cal P}^l$. Consider the simple network shown in Figure~\ref{fg.background_figure}(b) which has three hidden layers with two units in each of the hidden layers. In Figure~\ref{fg.background_figure}(c) on the right, we show the partition of the input space by the hyperplanes defined by the hidden layers. From the first hidden layer, we have two hyperplanes that divide the input space into four regions as shown by the red solid and red dotted lines. The hyperplanes defined by the second hidden layer depend on the active/inactive states of the units in the first hidden layer. As a result, the hyperplanes defined by the second hidden layer have different parameters in different regions defined by the first hidden layer. The blue solid and dotted line segments define the hyperplanes from the second hidden layer. Similarly, we observe that the hyperplanes from the third hidden layer recursively subdivide the regions formed in the previous layers. 
        
        Every input vector $\vx$ leads to an activation pattern $\mathcal{S}$, i.e., the $\vx$ to the \dnn results in the activations of hidden units as defined in $\mathcal{S}$. If we know the activation pattern, the mapping from the input $\vx$ to the output $\vy$ is linear. We follow the definition given in ~\cite{Serra2018}, for linear regions:
        
        \begin{defn}
            Given a Piece Wise Linear ({\sc PWL}) function $F:\mathbb{R}^{n_0}\rightarrow \mathbb{R}^{n_{L+1}}$ represented by \dnn, a linear region is the set of inputs that correspond to a same activation pattern in the DNN.
            \label{def:linear_regions2}
        \end{defn}
        
        In Figure~\ref{fg.background_figure}(c) we show activation patterns $(\{1,2\},\{1\},\{\})$ and $(\{1,2\},\{1,2\},\{1,2\})$ associated with the two shaded linear regions. Several recent papers have used the idea of activation patterns in the context of linear regions~\cite{Raghu2017,Montufar2017,Serra2018}, and show upper and lower bounds on the number of linear regions.

\section{MILP for identifying stably active and stably inactive units}\label{app.MILPforStableUnits}
    We identify which units are stable by running two MILP problems for each unit of the network, which gives us the maximum and the minimum output of each unit. If the minimum is non-negative, the unit is stably active. If the maximum is non-positive, the unit is stably inactive. 
    
    If the input space $\vx \in \mathbb{R}^{n_0}$ is bounded by minimum and maximum values along each dimension, which is typically the case in most imaging problems, then we define a MILP mapping polyhedral regions of $\vx$ to the output space $\mathbf{y}$. We use Boolean variables $z^l_i$ to denote if the unit $i$ in layer $l$ is active, or else if the complement is.
    
    For a given hidden unit $i$ in layer $l$ of the ReLU \dnn, the following set of constraints~\cite{serra2018empirical} maps the input to the output: 
    \begin{align}
        W_i^l h^{l-1} + \evb_i^l = \evg_i^l \label{eq:mip_unit_begin} \\
        \evg_i^l = \evh_i^l - \bar{\evh}_i^l  \label{eq:mip_after_Wb_begin} \\
        \evh_i^l \leq H_i^l \evz_i^l \\
        \bar{\evh}_i^l \leq \bar{H}_i^l (1-\evz_i^l) \\
        \evh_i^l \geq 0 \\
        \bar{\evh}_i^l \geq 0 \\
        \evz_i^l \in \{0, 1\} \label{eq:mip_unit_end}
    \end{align}
    
    The maximum bound on the individual units is calculated by solving a sequence of MILPs~\cite{serra2018empirical} on layers $l \in \{1,\dots,L\}$ for each unit $i \in n_{l}$ starting from the input layer  
    \begin{align}
        H_i^{l} = & ~\max ~ g_i^{l} \label{eq:h_ub_mip_begin} \\
        &\text{~~s.t.~~~} (\ref{eq:mip_unit_begin})-(\ref{eq:mip_unit_end}) &~\forall~l \in \{1, \dots, L\}, i \in \{1,\dots,n_{l}\}  \\
        & ~~~~~~~~~~~x \in D \label{eq:h_ub_mip_end}
    \end{align}
    The minimum bound is obtained by replacing (\ref{eq:h_ub_mip_begin}) with $\bar{H}_i^{l} = \max -g_i^{l}$.    
        
    MILP formulation (\ref{eq:h_ub_mip_begin}) - (\ref{eq:h_ub_mip_end}) is thus solved twice for every unit in the hidden layer. In case the optimal solution is found, we use the optimal solution as the bound. In case we do not get the solution, we use the optimal solution of the corresponding relaxed Linear Programming (LP) formulation as the bound which is generally available in most of the solvers for free. This is valid since the objective function of the MILP formulation is bounded by the corresponding relaxed LP formulation~\cite{conforti2014integer}.
    
    Because of the recursive nature of the formulation which involves all the units in all the layers before the current hidden layer, the formulation runs significantly slower in cases when there are multiple hidden layers in the network and when the actual bounds of the hidden units in the later hidden layers are estimated. 
    
    To achieve the computational speedup, these bounds of the units can be obtained approximately by first trying to obtain a single feasible solution greater than a threshold at each of the units. If one obtains a single feasible solution, it means that there exists some valid maximum and minima value for the units or the unit is definitely on and off. The actual bounds is later substituted by the objective value of the corresponding relaxed LP formulation. Although the approximation algorithm makes the formulation less tight, nevertheless it speeds up the computation for the bounds of the units in the later hidden layers. Also in case, we apply the approximation method and a feasible solution is obtained or even when the time limit is reached, we use the value of the bounds of the relaxed version of the MILP of the units as the optimal value.
    
    \paragraph{Adapting to classification models}
        The model of the MILP formulation (\ref{eq:h_ub_mip_begin}) - (\ref{eq:h_ub_mip_end}) requires that the each linear layer is followed by a ReLU. Typical classification network architectures such as LeNet \cite{LeCun1998} and also MLP architectures often employ softmax loss and therefore do not have the exact same architecture. The last layer of the classification network does not have the ReLU layer but instead is followed by softmax layer to get the probabilities. To adapt this formulation (\ref{eq:h_ub_mip_begin}) - (\ref{eq:h_ub_mip_end}) to modern neural networks, we do not estimate any bounds on the softmax layer. This is desirable since we do not want the units in this layer which predict the probabilities of the individual classes to be pruned.

\section{Enumeration of all linear regions}\label{app.MILPforlinRegions}
    All feasible activation patterns are first enumerated by solving the maximum and the minimum bounds of the individual units using the MILP formulation (\ref{eq:h_ub_mip_begin}) - (\ref{eq:h_ub_mip_end}) of~\cite{serra2018empirical}. Once those bounds have been known, we use the following single optimization problem of ~\cite{Serra2018}
    \begin{align}
        \max ~ & f \\
        \text{s.t.} ~ & (\ref{eq:mip_unit_begin})-(\ref{eq:mip_unit_end}) &~\forall~l \in \{1, \dots, L\}, i \in \{1,\dots,n_{l}\} \phantom{, H_i^l \ge 0}\\
        & f \leq \evh_i^l + (1 - \evz_i^l) H_i^l &~\forall~l \in \{1, \dots, L\}, i \in \{1,\dots,n_{l}\}, H_i^l \ge 0 \label{eq:mip_count_bound_f}\\
        & \vx \in \sX
    \end{align}
    It should be emphasized that the constraint (\ref{eq:mip_count_bound_f}) is only written for nodes which are stably active or are unstable. This constraint is not added when the unit is stably inactive since the values of $H_i^l$ is negative for such nodes and that makes the solution of the above optimization problem infeasible.

    The activation pattern is inferred by looking at the boolean variables $\evz_i^l$. To enumerate all the solution, we use the one-tree algorithm~\cite{Danna1}. In other words, once a solution is found, the solution is stored and removed from the solution set by using a Lazy cut where the goal of the lazy cut is to make the current obtained solution infeasible. We use the following simple constraint to remove the current combination of values of the binary variables $\evz$
    \begin{align}\label{eq:lazy_cut}
        \sum_{k=1}^{N_1} \evz_k - \sum_{i=1}^{N_0} \evz_k <= N_1 - 1
    \end{align}
    where $N_0$ and $N_1$ denote the number of binary variables of the hidden layers in the current solution that are $0$ and $1$ respectively.

\section{Proof of Theorem~\ref{th:1}}\label{app.proof:1}
    \begin{theorem-non}
        For a given neural network \dnnone, Algorithm~\ref{alg:compress} produces a neural network \dnntwo such that \dnnone and \dnntwo are local equivalent with respect to the input domain $D$. 
    \end{theorem-non}
    \begin{proof}
        If $H^l_i \leq 0$, then $h^l_i = 0$ for any input in $D$ and we consider unit $i$ in layer $l$ as stably inactive. Those units are analyzed by the block starting at line~\ref{lin:inactive}.  If there are other units left in the layer, which are either not stable or stably active but not removed, then removing unit $i$ does not affect the output of subsequent units since the output of the unit is always 0 in $D$. 
        If $\bar{H}^l_i \leq 0$, then $h_i^l = W^l_i h^{l-1} + b_i^l$ for any input in $D$ and we consider unit $i$ in layer $l$ as stably active. Those units are analyzed by the block starting at line~\ref{lin:active}. If the rank on the sub-matrix $W^l_A$ consisting of the weights of stably active units in set $A$ is the same as that of $W^l_{A \cup \{i\}}$, then $h^l_i = \sum\limits_{k \in A} \alpha_k w_k^l h^{l-1} + b_i^l = \sum\limits_{k \in A} \alpha_k (h^l_k - b^l_k) + b_i^l$. Therefore, as long as there are other units in the layer, we remove $h^l_i$ from the activation functions in layer $l+1$ by adding $\alpha_k w^{l+1}_{ji}$ to $w^{l+1}_{jk}$ and $w^{l+1}_{ji} \left( b_i^l - \sum\limits_{k \in A} \alpha_k b^l_k \right)$ to $b_j^{l+1}$. This trivially applies in cases where $w^l_i = $ \textbf{0}, even if $|A| = 0$.
        
        If all units left in layer $l$ are stably active, 
        then we remove the entire layer $l$ and connect layer $l-1$ directly with layer $l+1$, 
        since layer $l$ is equivalent to a linear transformation. 
        Since $h^l_k = W^l_j h^{l-1} + b^l_k$ for each stably active unit $k$ in layer $l$, 
        then $h^{l+1}_i = W^{l+1}_i h^l + b^{l+1}_i = W^{l+1}_i \left( \sum\limits_{k=1}^{n_l} W^l_k h^{l-1} + b^l_k \right) + b^{l+1}_i = \sum\limits_{j \in n_{l-1}} \left( \sum\limits_{k \in A} w^l_{kj} w^{l+1}_{ik} \right) h^{l-1}_j + b^{l+1}_i + \left( \sum\limits_{k \in A} w^{l+1}_{ik} b^l_k \right)$.
        
        If the only unit left in layer $l$ is stably inactive, then any input in $D$ results in $h^l =$ \textbf{0}, and consequently the neural network coincides with a constant function $f : x \rightarrow \Upsilon$. Therefore, we remove all hidden layers and replace the activation function of each output unit $i$ with a constant function mapping to $\Upsilon_i$. 
    \end{proof}

\section{Proof of Theorem~\ref{th:2}}\label{app.proof:2}
    \begin{theorem-non}\label{th:2_new}
        Given a \dnnone with associated function $\fone:\mathbb{R}^{n_0}\rightarrow \mathbb{R}^{n_{L+1}}$ with $L$ hidden layers and $n_l$ units in each layer $l$ (see Figure~\ref{fg.intro_figure}), there exists a global linear approximation to \dnntwo with associated function $\ftwo:\mathbb{R}^{n_0}\rightarrow \mathbb{R}^{n_{L+1}}$ such that, for a given norm $\ell$ and distance $\epsilon$, $\fone(\vx)=\ftwo(\vx)$ for any $\vx \in \mathbb{R}^{n_0}$ such that $\fone$ is a linear function on $\{ \vx' : \|\vx-\vx'\|_{\ell} \le \epsilon\}$. The transformed network \dnntwo can be constructed using only 2 hidden layers as shown in Figure~\ref{fg.general_case} with widths $(n'_1,n'_2)$ given by
        \begin{align}
            n'_1 &= \sum_{l=1}^{L-1} \left(2^{n_1+\dots+n_{l-1}}\right)2n_l + \left(2^{n_1+\dots+n_{L-1}}\right)n_L \label{eq.NoUnits1stLayer_1} \\\text{~and~~~~}
            n'_2 &= \left(2^{n_1+\dots+n_{L-1}}\right)n_{L+1} \label{eq.NoUnits2ndLayer_2}
        \end{align}
    \end{theorem-non}
    
    \begin{figure}[!htb]
        \centering
        \input{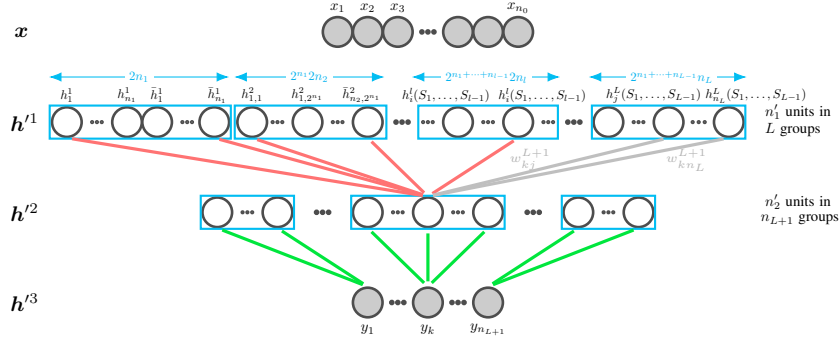}
        \caption{\emph{The transformed 2-hidden layer shallow network}. Each unit in $\vh'^{1}$ layer is connected to each of the input node with weights and biases given by the linear transformation till the layer in the original network. Each unit in $\vh^2$ layer is connected exactly to $\sum_{l=1}^{L}n_l$ units where $n_l$ denotes the number of units in layer $l$ of the original network. The blue rectangles denote the groups and the number of units in each of the group is shown in blue. The red line denotes the edge where a large negative weight $H$ is applied. The green line denote the edge where edge weight is $1$. The width of the network is made even more smaller by accounting only for the feasible activation patterns which are generally much less than theoretically possible patterns.}
    \end{figure}
    
    \begin{proof}
        We prove this result by constructing a 2-hidden layer network \dnntwo with associated function $\ftwo:\mathbb{R}^{n_0}\rightarrow \mathbb{R}^{n_{L+1}}$ that is a global approximation of the original $L$-layer \dnnone with associated function $\fone:\mathbb{R}^{n_0}\rightarrow \mathbb{R}^{n_{L+1}}$ for a given maximum offset parameter $\epsilon$. We use the notation $\vx \in R(S_1,\dots,S_{L-1})$, if the input point $\vx$ lies inside the linear region defined by the activation pattern $(S_1,\dots,S_{L-1})$. In other words, for every point $\vx$ that is at least $\epsilon$ distance from the boundary of any two linear regions,
        network \dnntwo is such that $\fone(\vx) = \ftwo(\vx)$.
        
        Each of the hidden units $h^l_i$ depends on the values of the hidden units in the previous layer and the weight parameters are given below:
        \begin{equation*}
            h^l_i  =  \max\left(0,w^{l}_{i1} h^{l-1}_1 + w^{l}_{i2} h^{l-1}_2 + \dots + w^{l}_{in_l} h^{l-1}_{n_l} + b^l_i\right) 
        \end{equation*}
                
        The value of $h^l_i$ is computed using the activations in the previous layer. By knowing the input vector $\vx$ and the activation sets of all the previous layers one could find the value of $h^l_i$. In other words, we could think of $h^l_i$ as a function that depends on the activation sets of all the previous layers and the input as $\vx$ as shown below:
        \begin{equation*}
            h^l_i  =  \max\left(0,g\left(\vx,S_1,\dots,S_{l-1}\right)\right)
        \end{equation*}
        where $g\left(\vx,S_1,\dots,S_{l-1}\right)$ is a linear function that depends only on the input vector $\vx$ and the activation sets of all the previous layers.
                
        We also introduce variables to denote units that are not activated as shown below:
        \begin{align*}
            \bar{h}^l_i &= \max\left(0,-\left(w^{l}_{i1} h^{l-1}_1 + w^{l}_{i2} h^{l-1}_2 + \dots + w^{l}_{in_l} h^{l-1}_{n_l} + b^l_i\right)\right) \\
            \bar{h}^l_i &= \max\left(0,-g\left(\vx,S_1,\dots,S_{l-1}\right)\right)
        \end{align*}
        
        Only one of the two variables $\{h^l_i,\bar{h}^l_i\}$ is non-negative given the input $\vx$ and activations of the previous $l-1$ layers. The basic idea in the transformation is shown in Figure~\ref{fg.general_case}: the first layer in the transformed network consists of an exponential number of units to allow for all possible activation patterns to occur, and additional nodes to enforce penalties for incorrect activations, i.e., all the activations other than the one that actually occur. In particular, the first hidden layer in the transformed shallow network consists of $L$ groups (shown in blue rectangles) totalling $n'_1$ hidden units as given in (~\ref{eq.NoUnits1stLayer}), where the first $L-1$ groups are used for imposing penalties for the activation patterns that do not occur, and the last group builds the necessary units for computing the underlying piece-wise linear function. The $l^\text{th}$ penalty group consists of $\left(2^{n_1+\dots+n_{l-1}}\right)2n_l$ units as given below:
        \begin{align*}
            &~ h^l_i(\vx,S_1,\dots,S_{l-1}),~\bar{h}^l_i(\vx,S_1,\dots,S_{l-1}),\\
            &~ ~~~~~~~~~~~~\forall~S_j \subseteq \{1,\dots,n_j\},~i \in \{1,\dots,n_l\}
        \end{align*}
        We get $\left(2^{n_1+\dots+n_{l-1}}\right)2n_l$ units because we have two units for ($h^l_i,\bar{h}^l_i$) for every combination of $S_j$ and $i$ in the above expression. 
    
        The last group consists of $2^{n_1+\dots+n_{L-1}}n_L$ units as given below: 
        \begin{equation}
            h^L_i\left(\vx,S_1,\dots,S_{L-1}\right),~i \in \{1,\dots,n_L\}
        \end{equation}
        
        Note that the units $h^L_i$ are exactly the same as the hidden units from the $L^{\text{th}}$ layer in the original network, and thus, we have the same notations. However, we consider all possible activation patterns, and compute the associated values for the hidden units $h^L_i(\vx,S_1,\dots,S_{L-1})$. Note that in the first hidden layer of the transformed network, we also compute some values for  $h^L_i(\vx,S_1,\dots,S_{L-1})$ when $\vx \notin R(S_1,\dots,S_{L-1})$. In the second hidden layer, we consider the combination of the hidden units along with penalty terms, so that the hidden units from activation patterns corresponding to linear regions not containing the input (i.e, $\vx \notin R(S_1,\dots,S_{L-1})$) are cancelled out. The second hidden layer of the transformed network consists of $n'_2=\left(2^{n_1+\dots+n_{L-1}}\right)n_{L+1}$ units in $n_{L+1}$ groups as given by (\ref{eq.NoUnits2ndLayer}). In the group with index $j \in \left\{1,\dots,n_{L+1}\right\}$, we have the following $2^{n_1+\dots+n_{L-1}}$ units with individual unit index $k \in \left\{1,\dots, 2^{n_1+\dots+n_{L-1}}\right\}$:
        \begin{align}
            h'^2_{jk}(\vx,S_1,\dots,S_{L-1}) 
            &= \max\Biggl(0, \sum\limits_{i=1}^{n_L} w_{ji}^{L+1}h^L_i(\vx,S_1,\dots,S_{L-1}) + b_j^{L+1} \nonumber \\ 
            &~ \phantom{max}- \sum\limits_{l=1}^{L-1}\sum\limits_{s \in S_l} H\bar{h}^l_s\left(\vx,S_1,\dots,S_{l-1}\right) - \sum\limits_{l=1}^{L-1}\sum\limits_{s \not\in S_l} H h ^l_s\left(\vx,S_1,\dots,S_{l-1}\right)\Biggr) \nonumber   \\
            &~ \phantom{maxmaxmax}~\forall~S_l \subseteq \left\{1,\dots,n_l\right\} 
        \end{align}
        The transformation is only approximate because we don't have $h'^2_{jk}\left(\vx,S_1,\dots,S_{L-1}\right)=0$ when $\vx \notin R\left(S_1,\dots,S_{L-1}\right)$. In such cases, we either have $\bar{h}^l_s\left(S_1,\dots,S_{l-1}\right) > 0$ ,when $s \in S_l$, or $h^l_s\left(S_1,\dots,S_{l-1}\right) > 0$, when $s \notin S_l$. Ideally, the large constant $H$ should make entire second term in the $\max$ operator become negative and thus making $h'^2_{jk}\left(\vx,S_1,\dots,S_{L-1}\right)=0$. However, when $\bar{h}^l_s\left(S_1,\dots,S_{l-1}\right)$ or $h^l_s\left(S_1,\dots,S_{l-1}\right)$ terms are infinitesimally small positive values, then the second term in the $\max$ operator could be a small positive value and thereby leading to $h'^2_{jk}\left(\vx,S_1,\dots,S_{L-1}\right) \neq 0$. To make the construction globally approximate, we choose a large constant value for $H$ such that the functions are equivalent by allowing an offset $\epsilon$ to the input $\vx$. In order to do this, we identify input $\vx$, associated linear region, and the offset point $\vx'$ where we have the largest violation with respect to the offset point (i.e., when $\vx \notin R\left(S_1,\dots,S_{L-1}\right)$) by setting $H=1$:
        \begin{align}
            \vx^{*},\vx'^{*},S_1^{*},\dots,S_{L-1}^{*}
            &= \argmax_{\vx,\vx',(S_1,\dots,S_{L-1})}\Biggl(0,\nonumber \\
            & \phantom{max}\min_{\vx',\|\vx-\vx'\|_{\ell} \le \epsilon}\biggl(\sum_{i=1}^{n_L} w_{ji}^{L+1}h^L_i\left(\vx',S_1,\dots,S_{L-1}\right) + b_j^{L+1} \nonumber \\
            &~ \phantom{max}- \sum\limits_{l=1}^{L-1}\sum\limits_{s \in S_l} \bar{h}^l_s\left(\vx',S_1,\dots,S_{l-1}\right) - \sum\limits_{l=1}^{L-1}\sum\limits_{s \not\in S_l} h ^l_s\left(\vx',S_1,\dots,S_{l-1}\right)\biggr)\Biggr),   \nonumber \\
            &~ \phantom{maxmaxmaxmax}~\forall~S_l \subseteq \left\{1,\dots,n_l\right\},~\vx \notin R\left(S_1,\dots,S_{L-1}\right) 
        \end{align}
        We fix the value of $H$ to be a large constant using the following expression:
        \begin{equation}
            H = \dfrac{
            \sum\limits_{i=1}^{n_L} w_{ji}^{L+1}h^L_i\left(\vx'^{*},S_1^{*},\dots,S_{L-1}^{*}\right) + b_j^{L+1}}{\sum\limits_{l=1}^{L-1}\sum\limits_{s \in S_l^{*}} \bar{h}^l_s\left(\vx'^{*},S_1^{*},\dots,S_{l-1}^{*}\right) + \sum\limits_{l=1}^{L-1}\sum\limits_{s \not\in S_l^{*}} h^l_s\left(\vx'^{*},S_1^{*},\dots,S_{l-1}^{*}\right)}
        \end{equation}
        This ensures that for all $\vx$ that is infinitesimally close to the boundary of a linear region, we always find a neighborhood point $\vx'$ where we have $\fone(\vx') = \ftwo(\vx')$. For all points $\vx$ that is not infinitesimally close to the boundaries of linear region we have $\fone(\vx) = \ftwo(\vx)$.
        
        The $j^{\text{th}}$ element of final $n_{L+1}$ dimensional output vector $\mathbf{y}$ is given as:
        \begin{equation}
            y_j(\vx) = {\eftwo}_{j}(\vx)  =  \sum\limits_{S_{l} \subseteq \{1,\dots,n_l\}} 
            h'^2_{jk}\left(\vx,S_1,\dots,S_{L-1}\right) \\
        \end{equation}
    \end{proof}

\section{Analysis of the Transformed Network obtained by Theorem~\ref{th:2}}\label{app.analysis_net_th2}
    We consider \dnnone with associated function $\fone:\mathbb{R}^{n_0}\rightarrow \mathbb{R}^{m}$ with $L$ hidden layers. For simplicity, we assume $n$ units in each of the hidden layer $l$. We use Theorem~\ref{th:2} to obtain a 2-layered globally approximate transformation \dnntwo with associated function $\ftwo:\mathbb{R}^{n_0}\rightarrow \mathbb{R}^{m}$. Then, the number of units in the first hidden layer of the transformed network is given by
    \begin{align*}
        n'^{1} &= 2n + 2^{n}2n + \dots + 2^{(L-2)n}2n + 2^{(L-1)n}n\\
               &= 2n \left[1 + 2^{n} + \dots + 2^{(L-2)n}\right] + 2^{(L-1)n}n\\
               &= 2n \left[\dfrac{2^{(L-1)n}-1}{2^n-1} \right] +  2^{(L-1)n}n\\
               &= \dfrac{n}{2^n-1} \left[2^{(L-1)n}2 - 2 + 2^{Ln} - 2^{(L-1)n}\right]\\
               &= \dfrac{n}{2^n-1} \left[2^{Ln} + 2^{(L-1)n} - 2\right]
    \end{align*}
    and the number of units in the second hidden layer of the transformed network is given by
    \begin{align*}
        n'^{2} &= 2^{(L-1)n}m
    \end{align*}
    
    The number of units as well as non-zero weights and biases possible in the original and the transformed networks are compared in Table~\ref{tab:nodes_edges}. Since, 
    each unit in the second hidden layer of the transformed network is connected to each input unit, therefore the number of non-zero weights or edges between the first hidden layer and the input layer is $\dfrac{n_0n}{2^n-1} \left[2^{Ln} + 2^{(L-1)n} - 2\right]$. The number of biases equals the number of nodes in the first hidden layer and is therefore equal to $\dfrac{n}{2^n-1} \left[2^{Ln} + 2^{(L-1)n} - 2\right]$ 
    Also, each unit in the second hidden layer of the transformed network is connected to exactly $nL$ units of the first hidden layer, the number of non-zero weights between the second and the first hidden layer is  $2^{(L-1)n}mnL$. Biases are required in the second hidden layer as well and therefore second hidden layer units of the transformed network contains $2^{(L-1)n}m$ bias terms. Now, the third layer only requires only one of the activation being selected for each of the $m$ outputs with weight $1$. So, the number of non-zero weights required for connecting the output layer with the second hidden layer is $2^{(L-1)n}m$. No bias is required.
    
    \begin{table}[!htb]
        \centering
        \begin{tabular}{ccc}
            \toprule
            \textbf{~} & \textbf{Original Network} & \textbf{Transformed Network} \\
            \midrule
            Hidden Units & $Ln$ & $\dfrac{n}{2^n-1} \left[2^{Ln} + 2^{(L-1)n} - 2\right] + 2^{(L-1)n}m$ \\[1em]
            Non-zero Weights  & $n_0n + (L-1)n^2$ & $\dfrac{n_0n}{2^n-1} \left[2^{Ln} + 2^{(L-1)n} - 2\right]\phantom{ + 2^{(L-1)n}+}$\\
                            & $+ nm$ & $\phantom{\dfrac{n_0n}{2^n-1-2-}} + 2^{(L-1)n}mnL + 2^{(L-1)n}m$\\[1em]
            Non-zero Biases & $Ln + m$ & $\dfrac{n}{2^n-1} \left[2^{Ln} + 2^{(L-1)n} - 2\right] + 2^{(L-1)n}m $\\
            \bottomrule
        \end{tabular}
        \caption{Comparison of the hidden units, non-zero weights and non-zero biases of the original network \dnnone~and the transformed shallow network~\dnntwo obtained by using Theorem~\ref{th:2}.}
        \label{tab:nodes_edges}
    \end{table}
    
\begin{figure}[!htb]
        \centering
        \includegraphics[width=0.6\columnwidth]{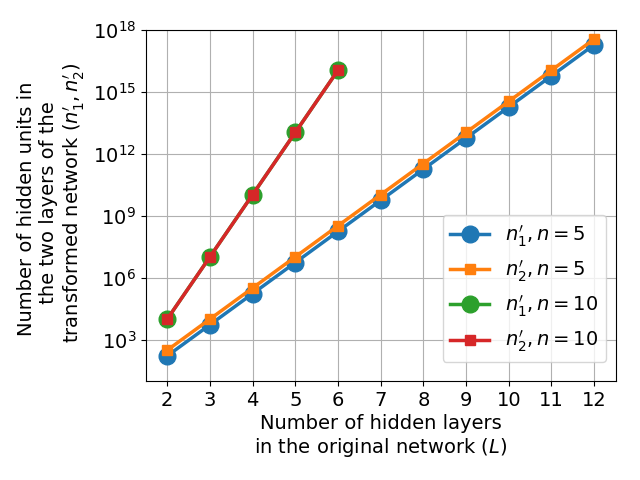}
        \caption{Variation in the number of hidden units of the first and second hidden layer of the transformed network ($n'_1$ and $n'_2$) of the transformed network with the number of hidden layers $(L)$ in the original deep neural network when the width of each hidden layer $(n)$ in the original network is fixed. Note that the y-axis is in log-scale.}
        \label{fig:vary_depth}
    \end{figure}

\begin{figure}[!htb]
        \centering
        \includegraphics[width=0.6\columnwidth]{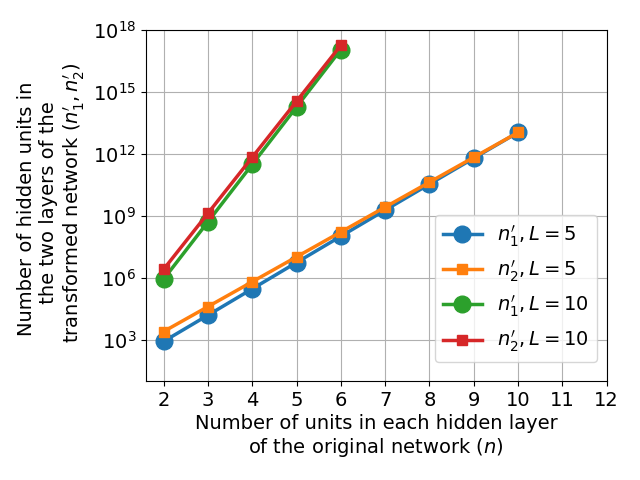}
        \caption{Variation in the number of hidden units of the first and second hidden layer of the transformed network $(n'_1$ and $n'_2)$ of the transformed network with the number of units in each hidden layer $(n)$ of the original deep neural network when the number of hidden layers $(L)$ in the original network is fixed. Note that the y-axis is in log-scale.}
        \label{fig:vary_width}
    \end{figure}

    Figures~\ref{fig:vary_depth} and~\ref{fig:vary_width} shows the variation in the number of hidden units of the first and second hidden layer of the transformed network ($n'_1$ and $n'_2$) of the transformed network with the number of hidden layers $(L)$ and 
    the number of units in each hidden layer $(n)$ in the original deep neural network keeping the width of each hidden layer $(n)$ and the number of hidden layers $(L)$ in the original network fixed respectively. We see linear plots in both the semilog figures which implies that the number of units required is indeed exponential as per Theorem~\ref{th:2}.
    

\section{Proof of Theorem~\ref{th:3}}\label{app.proof:3}
    \begin{theorem-non}
        Given a \dnnone~with associated function $\fone:\mathbb{R}^{n_0}\rightarrow \mathbb{R}^{n_{L+1}}$ with $L$ hidden layers and $n_l$ units in each layer $l$ (see Figure~\ref{fg.intro_figure}),  and with ${\cal A}$ denoting the set of all feasible activation patterns, there exists a globally approximate transformation \dnntwo with  associated function $\ftwo:\mathbb{R}^{n_0}\rightarrow \mathbb{R}^{n_{L+1}}$ such that, for a given norm $\ell$ and distance $\epsilon$, $\fone(\vx)=\ftwo(\vx)$ for any $\vx \in \mathbb{R}^{n_0}$ such that $\fone$ is a linear function on $\{ \vx' : \|\vx-\vx'\|_{\ell} \le \epsilon\}$. The transformed network \dnntwo can be constructed using only 2 hidden layers as shown in Figure~\ref{fg.intro_figure} with widths $(n'_1,n'_2)$ given by
        \begin{align*}
            n'_1 &= \sum\limits_{l=1}^{L-1} |{\cal A}_{l-1}|2n_l + |{\cal A}_{L-1}|n_L \\\text{~and~~~~} 
            n'_2 &= |{\cal A}_{L-1}|n_{L+1}
        \end{align*}
    
        The set ${\cal A}_l$ denotes the set of all subsets of feasible activation patterns till the $l$th layer, and any such subset $B \in {\cal A}_l$ is then given by 
        $    
        B = A - \underbrace{\{\emptyset,\dots,\emptyset,{\cal P}^{l+1},\dots,{\cal P}^{L}\}}_{L~sets},\forall~A \in {\cal A}
        $
    \end{theorem-non}
    
    \begin{proof}
        We prove this result by constructing a 2-hidden layer network \dnntwo  with associated function $\ftwo:\mathbb{R}^{n_0}\rightarrow \mathbb{R}^{n_{L+1}}$ that is a global approximation of the original $L$-layer \dnnone with associated function $\fone:\mathbb{R}^{n_0}\rightarrow \mathbb{R}^{n_{L+1}}$ for a given maximum offset parameter $\epsilon$. The transformed 2-layer network is similar to the one constructed by using Theorem~\ref{th:2} and shown in Figure~\ref{fg.general_case}, except that we discard all the hidden units in the transformed network that are redundant. In other words, the set of feasible activation patterns ${\cal A}$ allows us to discard all the units that correspond to infeasible activations.
        
        The second hidden layer in the transformed network \dnntwo consists of $L$ groups totalling $n'_1$ (as given in (\ref{eq.NoUnits1stLayer_ap})) hidden units, where the first $L-1$ groups are used for imposing penalties for the activation patterns that does not occur, and the last group builds the necessary units for computing the underlying piece-wise linear function. The $l^{\text{th}}$ penalty group consists of $|{\cal A}_{l-1}|2n_l$ units as given below:
        \begin{align*}
            &~ h^l_i\left(S_1,\dots,S_{l-1}\right),~\bar{h}^l_i\left(\vx,S_1,\dots,S_{l-1}\right),\\
            &~ ~~~~~~~~~~~~\forall~(S_1,\dots,S_{l-1}) \in {\cal A}_{l-1},~i \in \{1,\dots,n_l\}
        \end{align*}
        
        The last group consistd of $|{\cal A}_{L-1}|n_L$ units: 
        \begin{equation*}
            h^L_i\left(\vx,S_1,\dots,S_{L-1}\right),~i \in \{1,\dots,n_L\},~\left(S_1,\dots,S_{L-1}\right) \in {\cal A}_{L-1}
        \end{equation*}
        We have $|{\cal A}_{L-1}|n_L$ units by considering all indices for $i$ and all feasible subsets $(S_1,\dots,S_{L-1}) \in {\cal A}_{L-1}$.
        
        The second hidden layer consists of $n'_2= |{\cal A}_{L-1}|n_{L+1}$ units (as given in (\ref{eq.NoUnits2ndLayer_ap})) in $n_{L+1}$ groups. In the group with index $j \in \{1,\dots,n_{L+1}\}$, we have the following $|{\cal A}_{L-1}|$ units with individual unit index $k \in \left\{1,\dots,|{\cal A}_{L-1}|\right\}$:
        \begin{align}
            h'^2_{jk}(\vx,S_1,\dots,S_{L-1}) 
            &= \max\Biggl(0, \sum\limits_{i=1}^{n_L} w_{ji}^{L+1}h^L_i(\vx,S_1,\dots,S_{L-1}) + b_j^{L+1} \nonumber \\
            &~ \phantom{max}- \sum\limits_{l=1}^{L-1}\sum\limits_{s \in S_l} H\bar{h}^l_s\left(\vx,S_1,\dots,S_{l-1}\right) - \sum\limits_{l=1}^{L-1}\sum\limits_{s \not\in S_l} H h ^l_s\left(\vx,S_1,\dots,S_{l-1}\right)\Biggr), \nonumber \\
            &~ \phantom{maxmaxmax}~\forall~(S_1,\dots,S_{L-1}) \in {\cal A}_{L-1} 
        \end{align}
        
        The large constant $H$ is computed in the same manner as shown before in Appendix~\ref{app.proof:2} to ensure global approximate transformation with a given maximum offset $\epsilon$. 
        
        The $j^{\text{th}}$ element of final $n_{L+1}$ dimensional output vector $\mathbf{y}$ is given as:
        \begin{equation}
            y_j(\vx) = {\eftwo}_{j}(\vx)  =  \sum_{S_{l} \subseteq \{1,\dots,n_l\}} h'^2_{jk}(\vx,S_1,\dots,S_{L-1}) \\
        \end{equation}
    \end{proof}

\section{Effect of Regularization on Activation Patterns}\label{app.act_with_regularization}
    We trained the following different MLP architectures with varying number of layers and different number of hidden units in each of the hidden layers on MNIST. All these architectures used 784 input nodes and 10 output nodes with softmax as the loss function and ReLU as the activation function. We use a batch size of $64$ and SGD with a learning rate of $0.01$ and momentum of $0.9$ for training the model to $120$ epochs. The learning rate is decayed by a factor of $0.1$ after every $50$ epochs. The weights of the network are initialized with the Kaiming initialization \cite{He2016DeepRL} and the biases are initialized to zero. The training is carried out in Pytorch~\cite{paszke2017automatic}. The range for each of the input units is considered to be $[0, 1]$. We test out the effect of following different regularization on the number of activation patterns-
    \begin{itemize}
        \item Vanilla - No regularization added.
        \item $l_2$   - $l_2$ regularization weight of $0.003$.
        \item $l_1$   - $l_1$ regularization with weight $0.003$ on the first layer as in \cite{xiao2018training}.
        \item Dropout (DO) - Dropout of $0.3$ on the last layer.
        \item Dropout + $l_2$ regularization.
        \item Dropout + $l_1$ regularization.
        \item Adv + $l_1$ - Adversarial Training with PGD where PGD attacks are carried out with \lInfAttackD{0.15}{200}{0.1} and $l_1$ regularization as in \cite{xiao2018training}.
    \end{itemize}
    
    We did not test Adv alone since \cite{xiao2018training} shows that Adv coupled with $l_1$ works better. We also tested Adv+$l_1$+Dropout on one of the architectures. However, activation patterns obtained by Adv+$l_1$+Dropout was more than using Adv+$l_1$. Table~\ref{tab:regul_act_patterns} shows the number of activation patterns obtained by using different regularization. This shows the minimum number of activation patterns are obtained when models are trained with adversarial training and $l_1$ regularization whenever such models converge.
    
    \begin{table}[!htb]
        \centering
        \begin{tabular}{c!{\vrule width 1pt}ccccccc}
        \Xhline{3\arrayrulewidth}
        \textbf{Hidden} & \textbf{Vanilla} &\textbf{$\mathbf{l_2}$}& \textbf{$\mathbf{l_1}$} &\textbf{DO}& \textbf{DO+$\mathbf{l_2}$} & \textbf{DO+$\mathbf{l_1}$}  &\textbf{Adv+$\mathbf{l_1}$}\\
        \textbf{Architecture} &\gcellhead{SI}{VA$\mathbf{(\%)}$} &\gcellhead{SI}{VA$\mathbf{(\%)}$}&\gcellhead{SI}{VA$\mathbf{(\%)}$}&\gcellhead{SI}{VA$\mathbf{(\%)}$}&\gcellhead{SI}{VA$\mathbf{(\%)}$}&\gcellhead{SI}{VA$\mathbf{(\%)}$}&\gcellhead{SI}{VA$\mathbf{(\%)}$}\\
        \Xhline{2.5\arrayrulewidth}
        \multirow{2}{*}{$[$5,5,5,5$]$} & 2050             & 4569             & \textbf{458}               & 3548             & 2430             & 1411             & DNC\\
                                       &\gcell{1}{83.96} &\gcell{1}{87.59} &\gcell{3}{82.89} &\gcell{2}{79.51} &\gcell{2}{79.96} &\gcell{3}{70.61} &\gcell{-}{-}\\
        \Xhline{1.5\arrayrulewidth}
        \multirow{2}{*}{$[$5,5,10,10$]$} & 18350             & 8637             & \textbf{1696}               & 37000+             & 1871             & 27000             & DNC\\
                                       &\gcell{3}{82.87} &\gcell{3}{91.27} &\gcell{5}{82.79} &\gcell{0}{85.64} &\gcell{3}{89.22} &\gcell{2}{87.03} &\gcell{-}{-}\\
        \Xhline{1.5\arrayrulewidth}
        \multirow{2}{*}{$[$8,8,8,8$]$} & 108000+             & 103383             & 297000+               & 113000+             & 66000+             & 308000+             & \textbf{15137}\\
                                       &\gcell{0}{93.07} &\gcell{3}{93.81} &\gcell{1}{93.61} &\gcell{0}{90.71} &\gcell{1}{91.83} &\gcell{1}{90.05} &\gcell{4}{81.97}\\
        \Xhline{1.5\arrayrulewidth}
        \multirow{2}{*}{$[$5,5,5,5,5$]$} & 10548             & \textbf{1121}             & DNC               & 7909             & 2142             & DNC             & DNC\\
                                       &\gcell{1}{85.42} &\gcell{5}{87.55} &\gcell{-}{-} &\gcell{2}{62.95} &\gcell{4}{72.23} &\gcell{-}{-} &\gcell{-}{-}\\
        \Xhline{1.5\arrayrulewidth}
        \multirow{2}{*}{$[$5,5,5,10,10$]$} & 36912             & \textbf{1734}             & 15580               & 47256             & 5650             & 4911             & DNC\\
                                       &\gcell{1}{89.95} &\gcell{4}{88.42} & \gcell{4}{87.08} &\gcell{2}{84.42} &\gcell{4}{85.73} &\gcell{2}{76.69} &\gcell{-}{-}\\
        \Xhline{1.5\arrayrulewidth}
        \multirow{2}{*}{$[$8,8,8,8,8$]$} & 211000             & 403000             & 685000               & 700000+             & 106752             & 184572             & \textbf{447}\\
                                       &\gcell{1}{93.92} &\gcell{2}{94.03} &\gcell{2}{92.74} &\gcell{2}{91.70} &\gcell{4}{89.59} &\gcell{4}{88.03} &\gcell{12}{51.40}\\
        \Xhline{1.5\arrayrulewidth}
        \multirow{2}{*}{$[$5,5,5,5,5,5$]$}  & 21700             & \textbf{3986}             & DNC              & 6274             & 8740             & DNC             & DNC\\
                                       &\gcell{2}{89.01} &\gcell{3}{87.68} &\gcell{-}{-} &\gcell{3}{74.20} &\gcell{4}{55.66} &\gcell{-}{-} &\gcell{-}{-}\\
        \Xhline{1.5\arrayrulewidth}
        \multirow{2}{*}{$[$5,5,5,10,10,10$]$} & 12676             & 4452             & 149838               & 232479             & 68418             & 6581             & \textbf{354}\\
                                       &\gcell{1}{89.62} &\gcell{4}{87.50} &\gcell{1}{89.09} &\gcell{1}{82.39} &\gcell{4}{89.26} &\gcell{7}{77.72} &\gcell{17}{51.08}\\
        \Xhline{1.5\arrayrulewidth}
        \multirow{2}{*}{$[$8,8,8,8,8,8$]$} & 261000             & 218000             & 327000               & 223000             & 285000             & 230000             & \textbf{8157}\\
                                       &\gcell{1}{92.63} &\gcell{4}{93.77} &\gcell{3}{92.24} &\gcell{0}{90.50} &\gcell{2}{92.35} &\gcell{5}{88.57} &\gcell{7}{47.75}\\
        \Xhline{1.5\arrayrulewidth}
        \end{tabular}
        \caption{Number of activation patterns obtained by training with different regularization. SI shows the number of Stably Inactive nodes while VA shows the Validation Accuracy in $\%$. DNC - Did Not Converge. Vanilla - no regularization added, $l_1$- $l_1$ regularization, $l_2$ - $l_2$ regularization, DO - Dropout, Adv - Adversarial Training with PGD. + denotes the program did not terminate before it was stopped. The number of inputs and outputs in all these architectures is 784 and 10 respectively.}
        \label{tab:regul_act_patterns}
    \end{table}

\section{Local Stability Analysis}\label{app.local_stab}
\begin{figure}[!htb]
        \centering
        \begin{subfigure}{.33\textwidth}
          \centering
          \includegraphics[width=\linewidth,width=\linewidth]{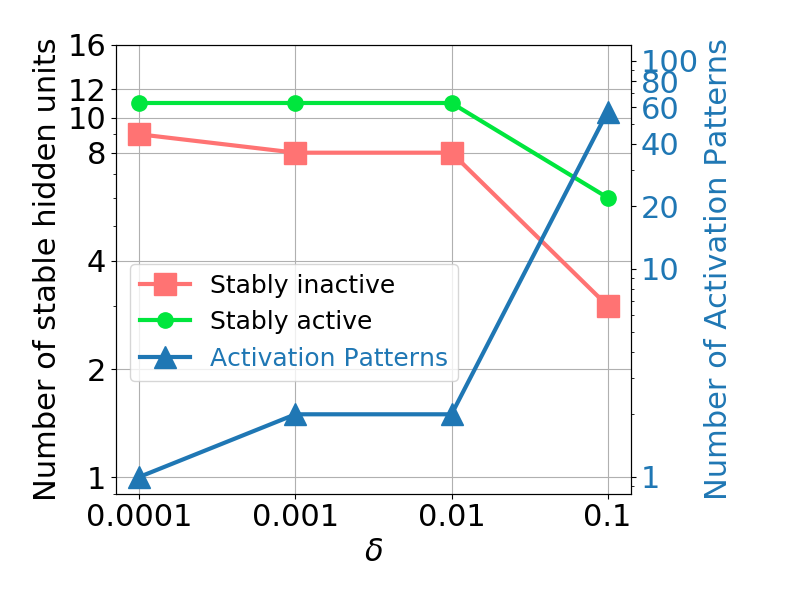}
          \caption{Class 0}
        \end{subfigure}%
        \begin{subfigure}{.33\textwidth}
          \centering
          \includegraphics[width=\linewidth,width=\linewidth]{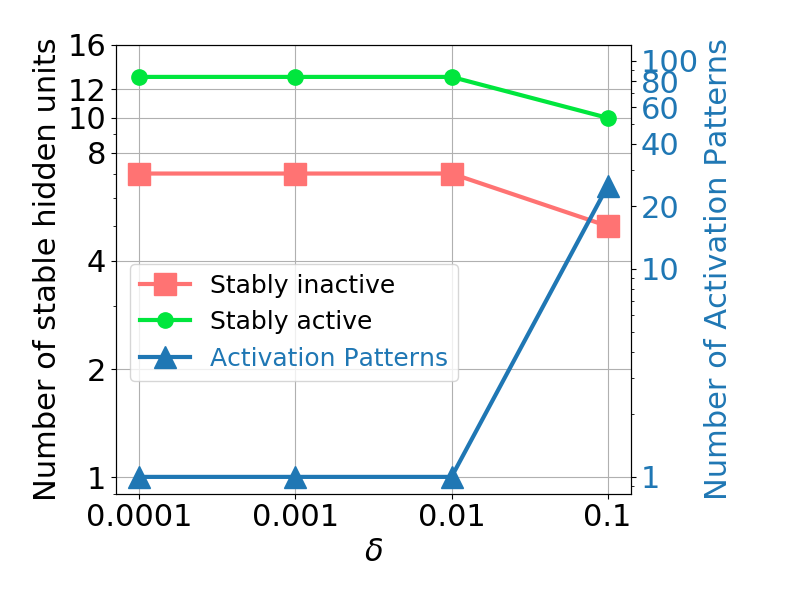}
          \caption{Class 1}
        \end{subfigure}%
        \begin{subfigure}{.33\textwidth}
          \centering
          \includegraphics[width=\linewidth,width=\linewidth]{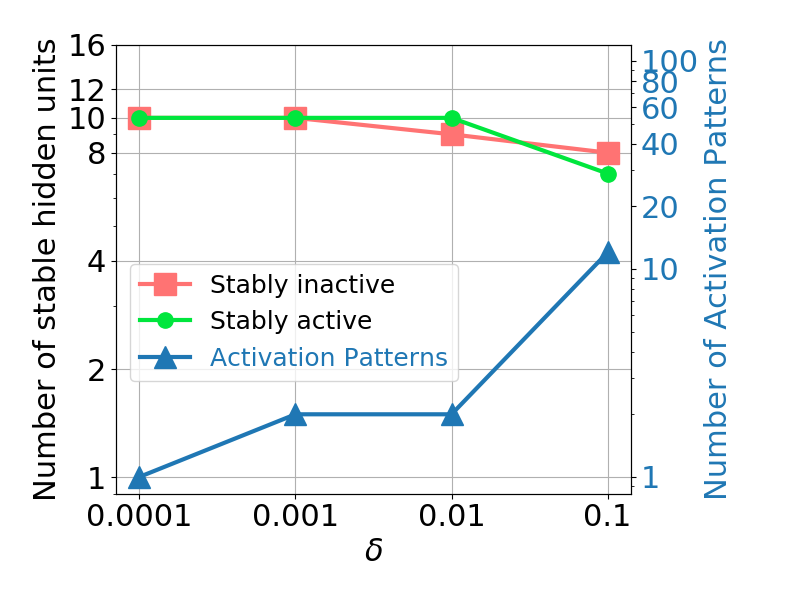}
          \caption{Class 2}
        \end{subfigure}%
        
        \medskip
        
        \begin{subfigure}{.33\textwidth}
          \centering
          \includegraphics[width=\linewidth,width=\linewidth]{figures/local_stability_class3.png}
          \caption{Class 3}
        \end{subfigure}%
        \begin{subfigure}{.33\textwidth}
          \centering
          \includegraphics[width=\linewidth,width=\linewidth]{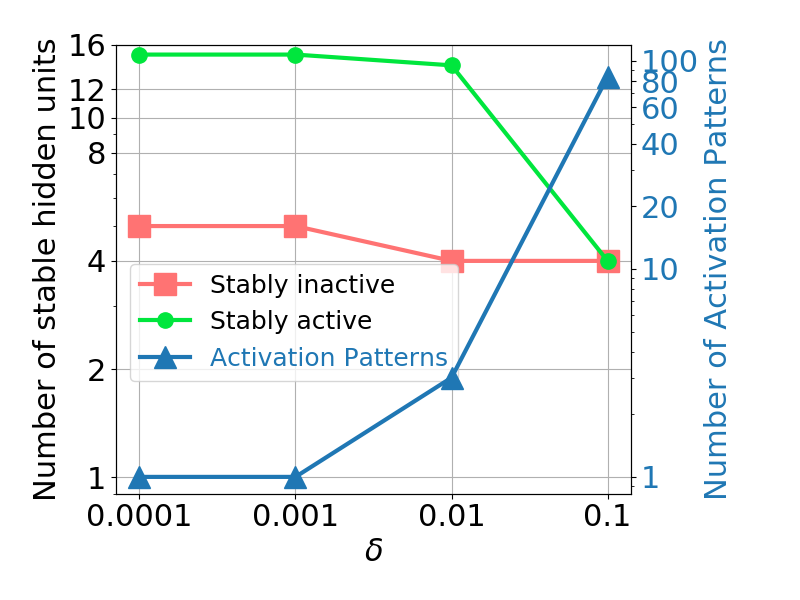}
          \caption{Class 4}
        \end{subfigure}%
        \begin{subfigure}{.33\textwidth}
          \centering
          \includegraphics[width=\linewidth,width=\linewidth]{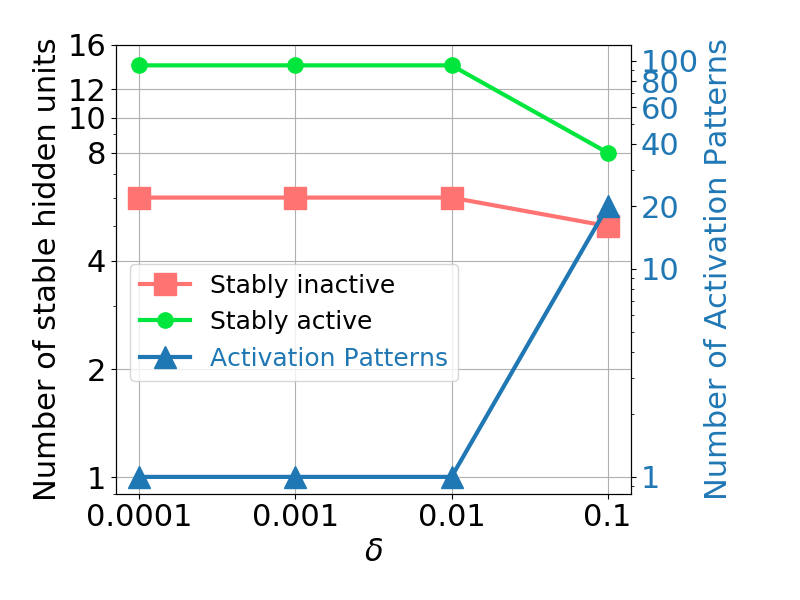}
          \caption{Class 5}
        \end{subfigure}%

        \medskip
        
        \begin{subfigure}{.33\textwidth}
          \centering
          \includegraphics[width=\linewidth,width=\linewidth]{figures/local_stability_class6.png}
          \caption{Class 6}
        \end{subfigure}%
        \begin{subfigure}{.33\textwidth}
          \centering
          \includegraphics[width=\linewidth,width=\linewidth]{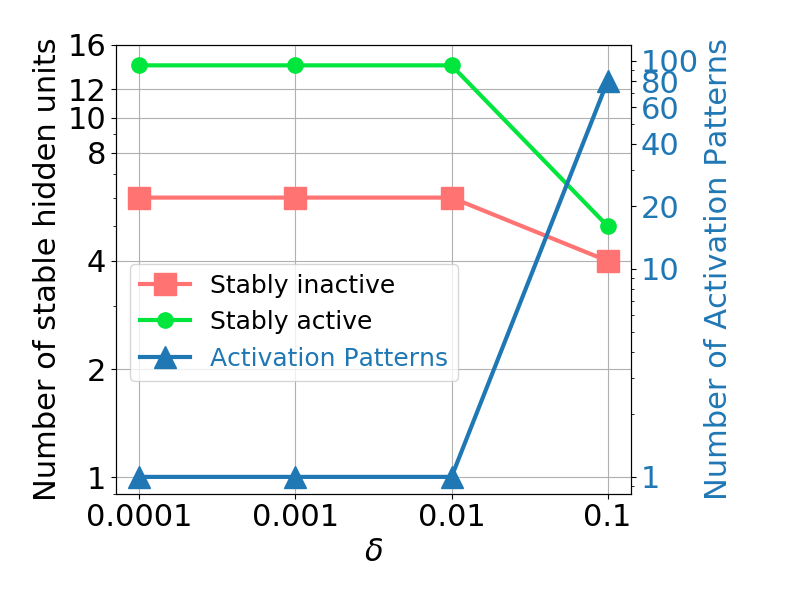}
          \caption{Class 7}
        \end{subfigure}%
        \begin{subfigure}{.33\textwidth}
          \centering
          \includegraphics[width=\linewidth,width=\linewidth]{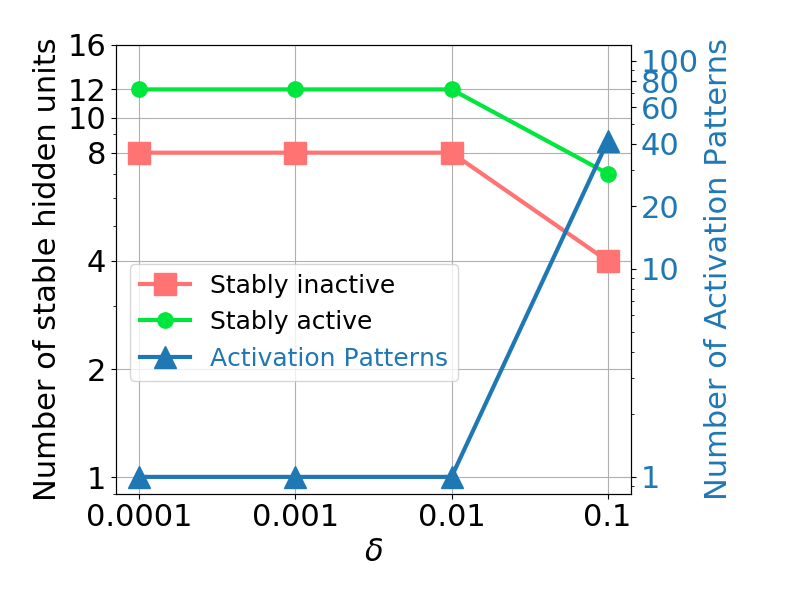}
          \caption{Class 8}
        \end{subfigure}%

        \medskip
        
        \begin{subfigure}{.33\textwidth}
          \centering
          \includegraphics[width=\linewidth,width=\linewidth]{figures/local_stability_class9.png}
          \caption{Class 9}
        \end{subfigure}%
        
        \caption{Variation in the number of stably inactive and stably active hidden units as well as activation patterns with $\delta$ for a randomly sampled example from each of the class.}
        \label{fig:loc_stab_vs_delta_2}
    \end{figure}

    Figure~\ref{fig:loc_stab_vs_delta_2} shows the variation of number of activation patterns and stably inactive hidden units with $\delta$ for a randomly sampled example from each of the class.

    \end{appendices}

\end{document}